\documentclass{article}

\usepackage{microtype}
\usepackage{graphicx}
\usepackage{subcaption}
\usepackage{booktabs} % for professional tables
\usepackage{amsmath}
\usepackage{amsfonts}
\usepackage{amssymb}
\usepackage{amsthm}
\usepackage{mathtools}
\usepackage{hyperref}

\usepackage[accepted]{icml2020}

\icmltitlerunning{Safe Imitation Learning via Fast Bayesian Reward Inference from Preferences}

\begin{document}

\twocolumn[
\icmltitle{Safe Imitation Learning via Fast Bayesian Reward Inference from Preferences}



\icmlsetsymbol{equal}{*}

\begin{icmlauthorlist}
\icmlauthor{Daniel S. Brown}{ut}
\icmlauthor{Russell Coleman}{ut,arl}
\icmlauthor{Ravi Srinivasan}{arl}
\icmlauthor{Scott Niekum}{ut}
\end{icmlauthorlist}

\icmlaffiliation{ut}{Computer Science Department, The University of Texas at Austin.}
\icmlaffiliation{arl}{Applied Research Laboratories, The University of Texas at Austin}

\icmlcorrespondingauthor{Daniel Brown}{dsbrown@cs.utexas.edu}
%\icmlcorrespondingauthor{Eee Pppp}{ep@eden.co.uk}

\icmlkeywords{Imitation Learning, Bayesian Reward Learning, Safety}

\vskip 0.3in
]



\printAffiliationsAndNotice{}  % leave blank if no need to mention equal contribution

\begin{abstract}
Bayesian reward learning from demonstrations enables rigorous safety and uncertainty analysis when performing imitation learning. However, Bayesian reward learning methods are typically computationally intractable for complex control problems. We propose Bayesian Reward Extrapolation (Bayesian REX), a highly efficient Bayesian reward learning algorithm that scales to high-dimensional imitation learning problems by pre-training a low-dimensional feature encoding via self-supervised tasks and then leveraging preferences over demonstrations to perform fast Bayesian inference. Bayesian REX can learn to play Atari games from demonstrations, without access to the game score and can generate 100,000 samples from the posterior over reward functions in only 5 minutes on a personal laptop. Bayesian REX also results in imitation learning performance that is competitive with or better than state-of-the-art methods that only learn point estimates of the reward function. Finally, Bayesian REX enables efficient high-confidence policy evaluation without having access to samples of the reward function. These high-confidence performance bounds can be used to rank the performance and risk of a variety of evaluation policies and provide a way to detect reward hacking behaviors.
\end{abstract}

\section{Introduction}
It is important that robots and other autonomous agents can safely learn from and adapt to a variety of human preferences and goals. One common way to learn preferences and goals is via imitation learning, in which an autonomous agent learns how to perform a task by observing demonstrations of the task \cite{Argall2009}. When learning from demonstrations, it is important for an agent to be able to provide high-confidence bounds on its performance with respect to the demonstrator; however, while there exists much work on high-confidence off-policy evaluation in the reinforcement learning (RL) setting, there has been much less work on high-confidence policy evaluation in the imitation learning setting, where the reward samples are unavailable. 

Prior work on high-confidence policy evaluation for imitation learning has used Bayesian inverse reinforcement learning (IRL) \cite{ramachandran2007bayesian} to allow an agent to reason about reward uncertainty and policy generalization error \cite{brown2018risk}. However, Bayesian IRL is typically intractable for complex problems due to the need to repeatedly solve an MDP in the inner loop, resulting in high computational cost as well as high sample cost if a model is not available. This precludes robust safety and uncertainty analysis for imitation learning in high-dimensional problems or in problems in which a model of the MDP is unavailable. We seek to remedy this problem by proposing and evaluating a method for safe and efficient Bayesian reward learning via preferences over demonstrations. Preferences over trajectories are intuitive for humans to provide \cite{akrour2011preference,wilson2012bayesian,sadigh2017active,christiano2017deep,palan2019learning} and enable better-than-demonstrator performance \cite{browngoo2019trex,brown2019drex}. To the best of our knowledge, we are the first to show that preferences over demonstrations enable both fast Bayesian reward learning in high-dimensional, visual control tasks as well as efficient high-confidence performance bounds.

We first formalize the problem of high-confidence policy evaluation \cite{thomas2015high} for imitation learning. We then propose a novel algorithm, Bayesian Reward Extrapolation (Bayesian REX), 
that uses a pairwise ranking likelihood to significantly increase the efficiency of generating samples from the posterior distribution over reward functions. We demonstrate that Bayesian REX can leverage neural network function approximation to learn useful reward features via self-supervised learning in order to efficiently perform deep Bayesian reward inference from visual demonstrations. Finally, we demonstrate that samples obtained from Bayesian REX can be used to solve the high-confidence policy evaluation problem for imitation learning. We evaluate our method on imitation learning for Atari games and demonstrate that we can efficiently compute high-confidence bounds on policy performance, without access to samples of the reward function. We use these high-confidence performance bounds to rank different evaluation policies according to their risk and expected return under the posterior distribution over the unknown ground-truth reward function. Finally, we provide evidence that bounds on uncertainty and risk provide a useful tool for detecting reward hacking/gaming \cite{amodei2016concrete}, a common problem in reward inference from demonstrations \cite{ibarz2018reward} as well as reinforcement learning \cite{ng1999policy,leike2017ai}.

\section{Related work}
\subsection{Imitation Learning}
%imitation learning and irl
Imitation learning is the problem of learning a policy from demonstrations and can roughly be divided into techniques that use behavioral cloning and techniques that use inverse reinforcement learning. Behavioral cloning methods \cite{pomerleau1991efficient,torabi2018behavioral} seek to solve the imitation learning problem via supervised learning, in which the goal is to learn a mapping from states to actions that mimics the demonstrator. While computationally efficient, these methods suffer from compounding errors \cite{ross2011reduction}. Methods such as DAgger \cite{ross2011reduction} and DART \cite{laskey2017dart} avoid this problem by repeatedly collecting additional state-action pairs from an expert.

Inverse reinforcement learning (IRL) methods seek to solve the imitation learning problem by estimating the reward function that the demonstrator is optimizing \cite{ng2000algorithms}.
% on the inferred reward function to learn a policy that can generalize well to states outside of the distribution of demonstrations. 
Classical approaches repeatedly alternate between a reward estimation step and a full policy optimization step \cite{abbeel2004apprenticeship,ziebart2008maximum,ramachandran2007bayesian}. Bayesian IRL \cite{ramachandran2007bayesian} samples from the posterior distribution over reward functions, whereas other methods seek a single reward function that induces the demonstrator's feature expectations \cite{abbeel2004apprenticeship}, often while also seeking to maximize the entropy of the resulting policy \cite{ziebart2008maximum}. 

Most deep learning approaches for IRL use maximum entropy policy optimization and divergence minimization between marginal state-action distributions \cite{ho2016generative,airl,ghasemipour2019divergence} and are related to Generative Adversarial Networks \cite{finn2016connection}. These methods scale to complex control problems by iterating between reward learning and policy learning steps. Alternatively, \citet{browngoo2019trex} use ranked demonstrations to learn a reward function via supervised learning without requiring an MDP solver or any inference time data collection. The learned reward function can then be used to optimize a potentially better-than-demonstrator policy. \citet{brown2019drex} automatically generate preferences over demonstrations via noise injection, allowing better-than-demonstrator performance even in the absence of explicit preference labels. However, despite their successes, deep learning approaches to IRL typically only return a point estimate of the reward function, precluding uncertainty and robustness analysis. % via a fully Bayesian approach. %One of our contributions is to propose the algorithm Bayesian Reward Extrapolation (Bayesian REX), the first deep Bayesian IRL algorithm that can scale to complex control problems with visual observations.

% Performing a single policy evaluation is a non-trivial task \cite{sutton2000policy} and even in tabular settings has complexity $O(N^3)$ where N is the size of the state-space \cite{littman1995complexity}. However, our proposed approach relies on thousands of policy evaluation steps. To make our approach tractable and to allow it to scale to high-dimensional control tasks such as Atari, we utilize self-supervised pre-training of reward feature to embed states into a pretrained latent space. We then perform MCMC sampling on this lower-dimensional space. This affords two main benefits: (1) MCMC remains the gold-standard for Bayesian Neural Networks, but is often impractical due to scaling issues. However, by only performing MCMC on the lower-dimensional latent space makes Bayesian inference tractable \cite{pradier2018projected}. (2) By learning a reward function that is a linear combination of learned features, policy evaluation can be performed via a simple dot product \cite{abbeel2004apprenticeship,barreto2017successor} reducing the complexity to be $O(w)$ where $w$ is the size of the latent reward feature space. 

\subsection{Safe Imitation Learning}
%Another contribution of this paper is an application of Bayesian REX to safe imitation learning.
While there has been much interest in imitation learning, less attention has been given to problems related to safety. SafeDAgger \cite{safedagger} and EnsembleDAgger \cite{menda2019ensembledagger} are extensions of DAgger that give control to the demonstrator in states where the imitation learning policy is predicted to have a large action difference from the demonstrator. %Control is given to the the demonstrator only if the predicted action difference of the novice is above some hand-tuned parameter, $\tau$. 
%Other work has focused making generative adversarial imitation learning (GAIL) \cite{ho2016generative} more robust and risk-sensitive. 
Other approaches to safe imitation learning seek to match the tail risk of the expert as well as find a policy that is indistinguishable from the demonstrations
\cite{majumdar2017risk,lacotte2019risk}.

\citet{brown2018efficient} propose a Bayesian sampling approach to provide explicit high-confidence performance bounds in the imitation learning setting, but require an MDP solver in the inner-loop. Their method uses samples from the posterior distribution $P(R|D)$ to compute sample efficient probabilistic upper bounds on the policy loss of any evaluation policy. 
%\citet{brown2018risk} extend this work by proposing an active learning algorithm that uses these high-confidence performance bounds for risk-aware policy improvement via active queries.
Other work considers robust policy optimization over a distribution of reward functions conditioned on demonstrations or a partially specified reward function, but these methods require an MDP solver in the inner loop, limiting their scalability \cite{hadfield2017inverse,brown2018risk,huang2018learning}. We extend and generalize the work of \citet{brown2018efficient} by demonstrating, for the first time, that high-confidence performance bounds can be efficiently obtained when performing imitation learning from high-dimensional visual demonstrations without requiring an MDP solver or model during reward inference. %We leave as future work the problem of robust policy optimization and improvement.

\subsection{Value Alignment and Active Preference Learning}
Safe imitation learning is closely related to the problem of value alignment, which seeks to design methods that prevent AI systems from acting in ways that violate human values \cite{hadfield2016cooperative,fisac2020pragmatic}. Research has shown that difficulties arise when an agent seeks to align its value with a human who is not perfectly rational \cite{milli2017should} and there are fundamental impossibility results regarding value alignment unless the objective is represented as a set of partially ordered preferences \cite{eckersley2018impossibility}.

Prior work has used active queries to perform Bayesian reward inference on low-dimensional, hand-crafted reward features \cite{sadigh2017active,brown2018risk,biyik2019asking}. \citet{christiano2017deep} and \citet{ibarz2018reward} use deep networks to scale active preference learning to high-dimensional tasks, but require large numbers of active queries during policy optimization and do not perform Bayesian reward inference. Our work complements and extends prior work by: (1) removing the requirement for active queries during reward inference or policy optimization, (2) showing that preferences over demonstrations enable efficient Bayesian reward inference in high-dimensional visual control tasks, and (3) providing  an efficient method for computing high-confidence bounds on the performance of any evaluation policy in the imitation learning setting.

%High-confidence off-policy evaluation
%phil and josiahs work for RL.
%borrow citations from my aaai paper

\subsection{Safe Reinforcement Learning}
Research on safe reinforcement learning (RL) usually focuses on safe exploration strategies or optimization objectives other than expected return \cite{garcia2015comprehensive}. Recently, objectives based on measures of risk such as value at risk (VaR) and conditional VaR have been shown to provide tractable and useful risk-sensitive measures of performance for MDPs \cite{tamar2015optimizing,chow2015risk}. Other work focuses on finding robust solutions to MDPs \cite{ghavamzadeh2016safe,petrik2019beyond}, using model-based RL to safely improve upon suboptimal demonstrations \cite{thananjeyan2019safety}, and obtaining high-confidence off-policy bounds on the performance of an evaluation policy \cite{thomas2015high,hanna2019importance}. %Recently, it has been shown that high-confidence off-policy evaluation is possible when samples of the true reward are available but the behavior policy is unknown \cite{hanna2019importance}. 
Our work %complements work high-confidence off-policy evaluation by 
provides an efficient solution to the problem of high-confidence policy evaluation in the imitation learning setting, in which samples of rewards are \textit{not observed} and the demonstrator's policy is unknown. 
%Try and cite papers from the invited speakers and organizers. 

%Bayesian neural network related lit.
\subsection{Bayesian Neural Networks}
%Bayesian Neural Networks seek to learn a posterior distribution over the weights of a neural network. Most applications involve supervised learning problems such as regression and classification, where the goal is to obtain a distribution on the outputs of a neural network. 
Bayesian neural networks typically either perform Markov Chain Monte Carlo (MCMC) sampling \cite{mackay1992practical}, variational inference \cite{sun2019functional,khan2018fast}, or use hybrid methods such as particle-based inference \cite{liu2016stein} to approximate the posterior distribution over neural network weights. Alternative approaches such as ensembles \cite{lakshminarayanan2017simple} or approximations such as Bayesian dropout \cite{gal2016dropout,kendall2017uncertainties} have also been used to obtain a distribution on the outputs of a neural network in order to provide uncertainty quantification \cite{maddox2019simple}. %In this work, %we are interested in the obtaining the posterior distribution over reward functions, conditioned on demonstrations and preferences. Thus, in theory any existing method for approximating the posterior distribution over weights in the reward function network could be used. However, 
We are not only interested in the uncertainty of the output of the reward function, but also in the uncertainty over the performance of a policy when evaluated under an uncertain reward function. 
Thus, we face the difficult problem of measuring the uncertainty in the evaluation of a policy, which depends on the stochasticity of the policy and the environment, as well as the uncertainty over the unobserved reward function.

\section{Preliminaries}

%\paragraph{Notation}
We model the environment as a Markov Decision Process (MDP) consisting of states $\mathcal{S}$, actions $\mathcal{A}$, transition dynamics $T:\mathcal{S} \times \mathcal{A} \times \mathcal{S} \to [0,1]$, reward function $R:\mathcal{S} \to \mathbb{R}$, initial state distribution $S_0$, and discount factor $\gamma$. 
%While we use a state-based reward function for the remainder of this paper, 
Our approach extends naturally to rewards defined as $R(s,a)$ or $R(s,a,s')$; however, state-based rewards have some advantages. \citet{airl} prove that a state-only reward function is a necessary and sufficient condition for a reward function that is disentangled from dynamics.
% and demonstrate that a state-only reward significantly improves imitation learning performance. 
Learning a state-based reward also allows the learned reward to be used as a potential function for reward shaping \cite{ng1999policy}, if a sparse ground-truth reward function is available.

A policy $\pi$ is a mapping from states to a probability distribution over actions. We denote the value of a policy $\pi$ under reward function $R$ as $V^\pi_R = \mathbb{E}_\pi[\sum_{t=0}^\infty \gamma^t R(s_t) | s_0 \sim S_0]$ and denote the value of executing policy $\pi$ starting at state $s \in \mathcal{S}$ as 
$V^\pi_R(s) = \mathbb{E}_\pi[\sum_{t=0}^\infty \gamma^t R(s_t) | s_0 = s]$. 
Given a reward function $R$, the Q-value of a state-action pair $(s,a)$ is $Q^{\pi}_R(s,a) = \mathbb{E}_\pi[\sum_{t=0}^\infty \gamma^t R(s_t) | s_0 = s, a_0 = a]$. We also denote $V^*_R = \max_{\pi} V^\pi_R$ and $Q^*_R(s,a) = \max_{\pi} Q^{\pi}_R(s,a)$.

%\paragraph{Bayesian Inverse Reinforcement Learning}
Bayesian inverse reinforcement learning (IRL) \cite{ramachandran2007bayesian} models the environment as an MDP$\setminus$R in which the reward function is unavailable. Bayesian IRL seeks to infer the latent reward function of a Boltzman-rational demonstrator that executes the following policy
\begin{equation} \label{eq:boltzman_demonstrator}
\pi^\beta_R(a|s) = \frac{e^{\beta Q^*_R(s,a)}}{\sum_{b \in \mathcal{A}} e^{\beta Q^*_R(s,b)}},
\end{equation}
in which $R$ is the true reward function of the demonstrator, and $\beta \in [0, \infty)$ represents the confidence that the demonstrator is acting optimally.
Under the assumption of Boltzman rationality, the likelihood of a set of demonstrated state-action pairs, 
$D = \{ (s,a) : (s,a) \sim \pi_D \}$,
given a specific reward function hypothesis $R$, can be written as 
\begin{equation} \label{eqn:boltzman_likelihood}
P(D | R) = \prod_{(s,a) \in D} \pi^\beta_R(a|s) = \prod_{(s,a) \in D} \frac{e^{\beta Q^*_R(s,a)}}{\sum_{b \in \mathcal{A}} e^{\beta Q^*_R(s,b)}}.
\end{equation}

Bayesian IRL generates samples from the posterior distribution
%\begin{equation}
$P(R|D) \sim P(D|R)P(R)$
%\end{equation}
via Markov Chain Monte Carlo (MCMC) sampling, but this requires solving for $Q^*_{R'}$ to compute the likelihood of each new proposal $R'$. Thus, Bayesian IRL methods are only used for low-dimensional problems with reward functions that are often linear combinations of a small number of hand-crafted features \cite{bobu2018learning,biyik2019asking}. One of our contributions is an efficient Bayesian reward inference algorithm that leverages preferences over demonstrations in order to significantly improve the efficiency of Bayesian reward inference.

\section{High Confidence Policy Evaluation for Imitation Learning} \label{sec:hcpe-il}
Before detailing our approach, we first formalize the problem of high-confidence policy evaluation for imitation learning. We assume access to an MDP$\setminus$R, an evaluation policy $\pi_{\rm eval}$, a set of demonstrations, $D = \{\tau_1,\ldots,\tau_m\}$, in which $\tau_i$ is either a complete or partial trajectory comprised of states or state-action pairs, a confidence level $\delta$, and performance statistic $g:\Pi \times \mathcal{R} \rightarrow \mathbb{R}$, in which $\mathcal{R}$ denotes the space of reward functions and $\Pi$ is the space of all policies.

The \textit{High-Confidence Policy Evaluation problem for Imitation Learning} (HCPE-IL) is to find a high-confidence lower bound $\hat{g}: \Pi \times \mathcal{D} \rightarrow \mathbb{R}$ such that
\begin{equation}
\text{Pr}(g(\pi_{\rm eval}, R^*) \geq \hat{g}(\pi_{\rm eval}, D)) \geq 1 - \delta,
\end{equation}
in which $R^*$ denotes the demonstrator's true reward function and $\mathcal{D}$ denotes the space of all possible demonstration sets. HCPE-IL takes as input an evaluation policy $\pi_{\rm eval}$, a set of demonstrations $D$, and a performance statistic, $g$, which evaluates a policy under a reward function. The goal of HCPE-IL is to return a high-confidence lower bound $\hat{g}$ on the performance statistic $g(\pi_{\rm eval}, R^*)$.

%Note that this problem setting is significantly more challenging than the standard high-confidence off-policy evaluation problem in reinforcement learning, which we denote as HCOPE-RL. In HCOPE-RL the behavior policy is typically known and the demonstrations from the behavior policy contain ground-truth reward samples \cite{thomas2015high,prasad2019defining}. In HCPE-IL, the behavior policy is the demonstrator's policy $\pi_{R^*}$, which is unknown, and must be estimated from demonstrations. Furthermore, in HCPE-IL the demonstrations contain only state-action pairs; samples of the true reward signal are not available. These key distinctions prevent us from using techniques such as importance sampling \cite{precup2000eligibility} which are commonly used to solve HCOPE-RL algorithms \cite{thomas2015high,hanna2019importance}. In the following sections we describe how to use preferences to scale Bayesian IRL to high-dimensional visual control tasks as a way to efficiently solve the HCPE-IL problem for complex, visual imitation learning tasks.

\section{Deep Bayesian Reward Extrapolation}
%As discussed in previously, Bayesian IRL, and methods that utilize it to obtain measures of uncertainty over reward functions, require repeatedly solving a policy optimization step, rendering them intractable for complex problems. This is because, for high-dimensional tasks such as learning control policies from pixel observations, even solving a single MDP can be challenging. Thus repeatedly sampling from $P(R|D)$ becomes intractable. 
%In this section we describe the main contribution of this paper: a method for scaling Bayesian reward function inference to complex, high-dimensional imitation learning problems.
We now describe our main contribution: a method for scaling Bayesian reward inference to high-dimensional visual control tasks as a way to efficiently solve the HCPE-IL problem for complex imitation learning tasks.
Our first insight is that the main bottleneck for standard Bayesian IRL \cite{ramachandran2007bayesian} is computing the likelihood function in Equation (\ref{eqn:boltzman_likelihood})
% \begin{equation}
% P(D | R) = \prod_{(s,a) \in D} \frac{e^{\beta Q^*_R(s,a)}}{\sum_{b \in \mathcal{A}} e^{\beta Q^*_R(s,b)}}.
% \end{equation}
which requires optimal Q-values. Thus, to make Bayesian reward inference scale to high-dimensional visual domains, it is necessary to either efficiently approximate optimal Q-values or to formulate a new likelihood. Value-based reinforcement learning focuses on efficiently learning optimal Q-values; however, for complex visual control tasks, RL algorithms can take several hours or even days to train \cite{mnih2015human,hessel2018rainbow}. This makes MCMC, which requires evaluating large numbers of likelihood ratios, infeasible given the current state-of-the-art in value-based RL. Methods such as transfer learning have great potential to reduce the time needed to calculate $Q^*_R$ for a new proposed reward function $R$; however, transfer learning is not guaranteed to speed up reinforcement learning \cite{taylor2009transfer}.
% and transfer learning methods that avoid performing reinforcement learning only provide loose bounds on the Q-values of a transferred policy's performance \cite{barreto2017successor}, which precludes accurate computation of the likelihood ratios needed for Bayesian inference \cite{ramachandran2007bayesian}. 
Thus, we choose to focus on reformulating the likelihood function as a way to speed up Bayesian reward inference. 

An ideal likelihood function requires little computation and minimal interaction with the environment. To accomplish this, we leverage recent work on learning control policies from preferences \cite{christiano2017deep,palan2019learning,biyik2019asking}. Given ranked demonstrations, \citet{browngoo2019trex} propose Trajectory-ranked Reward Extrapolation (T-REX): an efficient reward inference algorithm that transforms reward function learning into classification problem via a pairwise ranking loss. T-REX removes the need to repeatedly sample from or partially solve an MDP in the inner loop, allowing it to scale to visual imitation learning domains such as Atari and to extrapolate beyond the performance of the best demonstration. However, T-REX only solves for a point estimate of the reward function. We now discuss how a similar approach based on a pairwise preference likelihood allows for efficient sampling from the posterior distribution over reward functions.

We assume access to a sequence of $m$ trajectories, $D = \{ \tau_1,\ldots,\tau_m \}$, along with a set of pairwise preferences over trajectories $\mathcal{P} =  \{(i,j) : \tau_i \prec \tau_j \}$. Note that we do not require a total-ordering over trajectories. These preferences may come from a human demonstrator or could be automatically generated by watching a learner improve at a task \cite{jacq2019learning,browngoo2019trex} or via noise injection \cite{brown2019drex}. 
%Some trajectory pairs may not have preference information and some trajectory pairs may be equally preferred, i.e., $(i,j)$ and $(j,i)$ are both in the set $\mathcal{P}$. 
Given trajectory preferences, we can formulate a pair-wise ranking likelihood to compute the likelihood of a set of preferences over demonstrations $\mathcal{P}$, given a parameterized reward function hypothesis $R_\theta$. We use the standard Bradley-Terry model \cite{bradley1952rank}
%, alternatively called Luce's choice axiom \cite{luce2012individual}, 
to obtain the following pairwise ranking likelihood function, commonly used in learning to rank applications such collaborative filtering \cite{volkovs2014new}:  %\cite{cao2007learning,szummer2011semi,volkovs2014new}: 
\begin{equation}\label{eqn:pairwiserank}
P(D, \mathcal{P}   \mid R_\theta) = \prod_{(i,j) \in \mathcal{P}} \frac{e^{\beta R_\theta(\tau_j)}}{e^{\beta R_\theta(\tau_i)} + e^{\beta R_\theta(\tau_j)}},
\end{equation}
in which $R_\theta(\tau) = \sum_{s \in \tau} R_\theta(s)$ is the predicted return of trajectory $\tau$ under the reward function $R_\theta$, and $\beta$ is the inverse temperature parameter that models the confidence in the preference labels. We can then perform Bayesian inference via MCMC to obtain samples from $P(R_\theta \mid D, \mathcal{P}) \propto P(D, \mathcal{P}  \mid R_\theta) P(R_\theta)$. We call this approach Bayesian Reward Extrapolation or Bayesian REX.

Note that using the likelihood function defined in Equation (\ref{eqn:pairwiserank}) does not require solving an MDP. In fact, it does not require any rollouts or access to the MDP. All that is required is that we first calculate the return of each trajectory under $R_\theta$ and compare the relative predicted returns to the preference labels to determine the likelihood of the demonstrations under the reward hypothesis $R_\theta$. Thus, given preferences over demonstrations, Bayesian REX is significantly more efficient than standard Bayesian IRL. In the following section, we discuss further optimizations that improve the efficiency of Bayesian REX and make it more amenable to our end goal of high-confidence policy evaluation bounds.

%This likelihood function Given this preference-based likelihood function we can perform preference-based Bayesian reward learning using standard MCMC.% as described in Algorithm~\ref{alg:P-BIRL} in the Appendix to generate samples from $P(R \mid \mathcal{P},D) \sim P(D, \mathcal{P} \mid R) P(R)$. %When $R_\theta$ is a deep network we call this Deep Preference-based Bayesian IRL or DeeP-BIRL.

\begin{figure*}
\centering
\includegraphics[width=\linewidth]{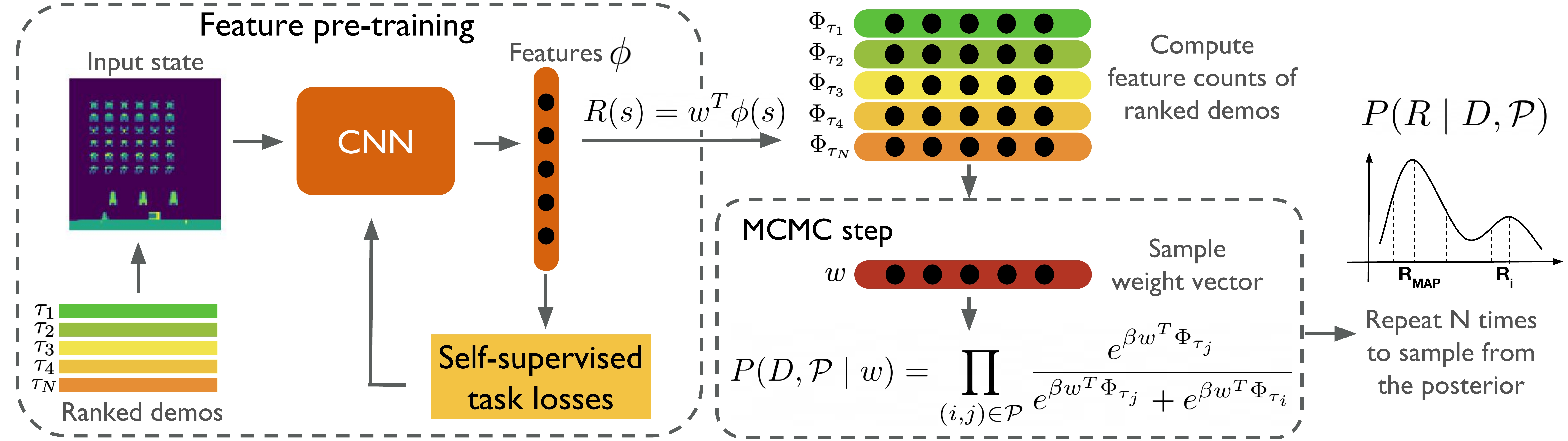}
\caption{Bayesian Reward Extrapolation uses ranked demonstrations to pre-train a low-dimensional state feature embedding $\phi(s)$ via self-supervised losses. After pre-training, the latent embedding function $\phi(s)$ is frozen and the reward function is represented as a linear combination of the learned features: R(s) = $w^T \phi(s)$. MCMC proposal evaluations use a pairwise ranking likelihood that gives the likelihood of the preferences $\mathcal{P}$ over demonstrations $D$, given a proposal $w$. By pre-computing the embeddings of the ranked demonstrations, $\Phi_{\tau_i}$, MCMC sampling is highly efficient---it does not require access to an MDP solver or data collection during inference.}
\label{fig:BayesianREX}
\end{figure*}

\subsection{Optimizations} \label{sec:optimizations}
In order to learn rich, complex reward functions, it is desirable to use a deep network to represent the reward function $R_\theta$. While MCMC remains the gold-standard for Bayesian Neural Networks, it is often challenging to scale to deep networks. To make Bayesian REX more efficient and practical, we propose to limit the proposal to only change the last layer of weights in $R_\theta$ when generating MCMC proposals---we will discuss pre-training the bottom layers of $R_\theta$ in the next section. After pre-training, we freeze all but the last layer of weights and use the activations of the penultimate layer as the latent reward features $\phi(s) \in \mathbb{R}^k$. This allows the reward at a state to be represented as a linear combination of $k$ features: $R_\theta(s) = w^T \phi(s)$. Similar to work by \citet{pradier2018projected}, operating in a lower-dimensional latent space makes full Bayesian inference tractable.

A second advantage of using a learned linear reward function is that it allows us to efficiently compute likelihood ratios when performing MCMC. Consider the likelihood function in Equation~(\ref{eqn:pairwiserank}). If we do not represent $R_\theta$ as a linear combination of pretrained features, and instead let any parameter in $R_\theta$ change during each proposal, then for $m$ demonstrations of length $T$, computing $P(D, \mathcal{P} \mid R_\theta)$ for a new proposal $R_\theta$ requires $O(mT)$ forward passes through the entire network to compute $R_\theta(\tau_i)$. Thus, the complexity of generating $N$ samples from the posterior results is $O(mTN |R_\theta|)$, where $|R_\theta|$ is the number of computations required for a full forward pass through the entire network $R_\theta$.  Given that we would like to use a deep network to parameterize $R_\theta$ and generate thousands of samples from the posterior distribution over $R_\theta$, this many computations will significantly slow down MCMC proposal evaluation. 

If we represent $R_\theta$ as a linear combination of pre-trained features, we can reduce this computational cost because 
\begin{equation}
R_\theta (\tau) = \sum_{s \in \tau} w^T\phi(s) = w^T\sum_{s \in \tau} \phi(s) = w^T \Phi_{\tau}.
\end{equation}
Thus, we can precompute and cache $\Phi_{\tau_i} = \sum_{s \in \tau_i} \phi(s)$ for $i = 1,\ldots,m$ and rewrite the likelihood as
\begin{equation}\label{eqn:lincombo_Boltzman}
P(D, \mathcal{P}  \mid R_\theta) = \prod_{(i,j) \in \mathcal{P}} \frac{e^{\beta w^T \Phi_{\tau_j}}}{e^{\beta w^T \Phi_{\tau_j}} + e^{\beta w^T \Phi_{\tau_i}}}.
\end{equation}
Note that demonstrations only need to be passed through the reward network once to compute $\Phi_{\tau_i}$ since the pre-trained embedding remains constant during MCMC proposal generation. This results in an initial $O(mT)$ passes through all but the last layer of $R_\theta$ to obtain $\Phi_{\tau_i}$, for $i=1,\ldots,m$, and then only $O(mk)$ multiplications per proposal evaluation thereafter---each proposal requires that we compute $w^T\Phi_{\tau_i}$ for $i=1,\ldots,m$ and $\Phi_{\tau_i} \in \mathbb{R}^k$. Thus, when using feature pre-training, the total complexity is only $O(mT|R_\theta| + mkN)$ to generate $N$ samples via MCMC. This reduction in the complexity of MCMC from $O(mTN |R_\theta|)$ to $O(mT|R_\theta| + mkN)$ results in significant and practical computational savings because (1) we want to make $N$ and $R_\theta$ large and (2) the number of demonstrations, $m$, and the size of the latent embedding, $k$, are typically several orders of magnitude smaller than $N$ and $|R_\theta|$.

%There are three advantages to this formulation: (1) the proposal dimension for MCMC is significantly reduced, allowing for faster convergence; (2) we can efficiently compute the expected value of a policy via a single dot product, and (3) the computation required to calculated the likelihood of a proposed MCMC sample is significantly reduced.

A third, and critical advantage of using a learned linear reward function is that it makes solving the HCPE-IL problem discussed in Section~\ref{sec:hcpe-il} tractable.
Performing a single policy evaluation is a non-trivial task \cite{sutton2000policy} and even in tabular settings has complexity $O(|S|^3)$ in which $|S|$ is the size of the state-space \cite{littman1995complexity}. Because we are in an imitation learning setting, we would like to be able to efficiently evaluate any given policy across the posterior distribution over reward functions found via Bayesian REX. Given a posterior distribution over $N$ reward function hypotheses we would need to solve $N$ policy evaluations. However, note that given $R(s) = w^T \phi(s)$, the value function of a policy can be written as \begin{equation}
V^\pi_R = \mathbb{E}_{\pi}[\sum_{t=0}^T R(s_t)] =  w^T \mathbb{E}_{\pi}[\sum_{t=0}^T \phi(s_t)] = w^T \Phi_\pi,
\end{equation}
in which we assume a finite horizon MDP with horizon $T$ and in which $\Phi_\pi$ are the expected feature counts \cite{abbeel2004apprenticeship,barreto2017successor} of $\pi$. Thus, given any evaluation policy $\pi_{\rm eval}$, we only need to solve one policy evaluation problem to compute $\Phi_{\rm eval}$. We can then compute the expected value of $\pi_{\rm eval}$ over the entire posterior distribution of reward functions via a single matrix vector multiplication $W \Phi_{\pi_{\rm eval}}$, where $W$ is an $N$-by-$k$ matrix with each row corresponding to a single reward function weight hypothesis $w^T$. This significantly reduces the complexity of policy evaluation over the reward function posterior distribution from $O(N |S|^3)$ to $O(|S|^3 + Nk)$. %Because policy evaluation can be performed via a simple dot product, the computational complexity of computing $M$ policy evaluations becomes $O(|S|^3 + M\cdot k)$ in which $k \ll |S|$ is the size of the latent reward feature space. In comparison, the complexity of policy evaluation without a linear reward function would be $O(M \cdot |S|^3)$. If we estimate the expected feature counts of a policy via $C$ Monte Carlo rollouts of length $T$, then the complexity of our approach is only $O(NT + Mk)$ as opposed to $O(MNT)$ without a linear reward function \textbf{[Does this make sense for monte carlo rollouts?]}.

When we refer to Bayesian REX we will refer to the optimized version described in this section (see the Appendix for full implementation details and pseudo-code)\footnote{Project page, code, and demonstration data are available at \url{https://sites.google.com/view/bayesianrex/}} .
%Given the expected feature counts $\Phi_{\pi_{\rm eval}}$ of an evaluation policy, 
Running MCMC with 66 preference labels to generate 100,000 reward hypothesis for Atari imitation learning tasks takes approximately 5 minutes on a Dell Inspiron 5577 personal laptop with an Intel i7-7700 processor without using the GPU. 
%The time to perform a Monte Carlo estimate of the expected feature counts of a policy takes \textbf{[time policy eval]} and only has to be computed once per evaluation policy. 
In comparison, using standard Bayesian IRL to generate \textit{one sample} from the posterior takes 10+ hours of training for a parallelized PPO reinforcement learning agent \cite{baselines} on an NVIDIA TITAN V GPU.

\subsection{Pre-training the Reward Function Network} \label{sec:pretraining}
The previous section presupposed access to a pretrained latent embedding function $\phi: S \rightarrow \mathbb{R}^k$. We now discuss our pre-training process. 
%If we had access to the reward, we could simply learn $\phi(s)$ via supervised learning on the reward samples; however, w
Because we are interested in imitation learning problems, we need to be able to train $\phi(s)$ from the demonstrations without access to the ground-truth reward function. 
One potential method is to train $R_\theta$ using the pairwise ranking likelihood function in Equation~(\ref{eqn:pairwiserank}) and then freeze all but the last layer of weights; however, the learned embedding may overfit to the limited number of preferences over demonstrations and fail to capture features relevant to the ground-truth reward function. Thus, we supplement the pairwise ranking objective with auxiliary objectives that can be optimized in a self-supervised fashion using data from the demonstrations. 

\begin{table}[t]
  \caption{Self-supervised learning objectives used to pre-train $\phi(s)$.}
  \label{tab:aux_losses}
  \centering
    \vspace{0.1cm}
\begin{tabular}{ll}
\toprule
%Pairwise Ranking &  $f(\phi(\tau_i), \phi(\tau_j)) \rightarrow \mathbf{1}_{\tau_i \succ \tau_j} $ \\
 Inverse Dynamics & $f_{\rm ID}(\phi(s_t), \phi(s_{t+1})) \rightarrow a_t$ \\
 Forward Dynamics & $f_{\rm FD}(\phi(s_t), a_t) \rightarrow s_{t+1}$ \\
 Temporal Distance & $f_{\rm TD}(\phi(s_t), \phi(s_{t+x}) \rightarrow x$ \\
 Variational Autoencoder & $f_{A}(\phi(s_t)) \rightarrow s_t$ \\
\bottomrule
\end{tabular}
\end{table}

We use the following self-supervised tasks to pre-train $R_\theta$: (1) Learn an inverse dynamics model that uses embeddings $\phi(s_t)$ and $\phi(s_{t+1})$ to predict the corresponding action $a_t$ \cite{torabi2018behavioral,hanna2017grounded}, (2) Learn a forward dynamics model that predicts $s_{t+1}$ from $\phi(s_t)$ and $a_t$ \cite{oh2015action,thananjeyan2019safety}, (3) Learn an embedding $\phi(s)$ that predicts the temporal distance between two randomly chosen states from the same demonstration \cite{imitationyoutube}, and (4) Train a variational pixel-to-pixel autoencoder in which $\phi(s)$ is the learned latent encoding \cite{makhzani2017pixelgan,doersch2016tutorial}. Table~\ref{tab:aux_losses} summarizes the self-supervised tasks used to train $\phi(s)$. 

There are many possibilities for pre-training $\phi(s)$. We used the objectives described above to encourage the embedding to encode different features. For example, an accurate inverse dynamics model can be learned by only attending to the movement of the agent. Learning forward dynamics supplements this by requiring $\phi(s)$ to encode information about short-term changes to the environment. Learning to predict the temporal distance between states in a trajectory forces $\phi(s)$ to encode long-term progress. Finally, the autoencoder loss acts as a regularizer to the other losses as it seeks to embed all aspects of the state (see the Appendix for details).
The Bayesian REX pipeline for sampling from the reward function posterior is shown in Figure~\ref{fig:BayesianREX}.

\subsection{HCPE-IL via Bayesian REX}
We now discuss how to use Bayesian REX to find an efficient solution to the high-confidence policy evaluation for imitation learning (HCPE-IL) problem (see Section~\ref{sec:hcpe-il}). 
Given samples from the distribution $P(w \mid D, \mathcal{P})$, where $R(s) = w^T \phi(s)$, we compute the posterior distribution over any performance statistic $g(\pi_{\rm eval}, R^*)$ as follows. For each sampled weight vector $w$ produced by Bayesian REX, we compute $g(\pi_{\rm eval}, w)$. This results in a sample from the posterior distribution $P(g(\pi_{\rm eval}, R) \mid D, \mathcal{P})$, i.e., the posterior distribution over performance statistic $g$. %conditioned on $D$ and $\mathcal{P}$.  
We then compute a $(1-\delta)$ confidence lower bound, $\hat{g}(\pi_{\rm eval}, D)$, by finding the $\delta$-quantile of $g(\pi_{\rm eval}, w)$ for $w \sim P(w\mid  D, \mathcal{P})$. 

While there are many potential performance statistics $g$, we chose to focus on bounding the expected value of the evaluation policy, i.e., $g(\pi_{\rm eval}, R^*) = V^{\pi_{\rm eval}}_{R^*} = {w^*}^T \Phi_{\pi_{\rm eval}}$. To compute a $1 - \delta$ confidence bound on $V^{\pi_{\rm eval}}_{R^*}$, we take advantage of the learned linear reward representation to efficiently calculate the posterior distribution over policy returns given preferences and demonstrations. This distribution over returns is calculated via a matrix vector product,  $W \Phi_{\pi_{\rm eval}}$,
in which each row of $W$ is a sample, $w$, from the MCMC chain and $\pi_{\rm eval}$ is the evaluation policy. We then sort the resulting vector and select the $\delta$-quantile lowest value. This results in a $1-\delta$ confidence lower bound on $V^{\pi_{\rm eval}}_{R^*}$ and corresponds to the $\delta$-Value at Risk (VaR) over $V^{\pi_{\rm eval}}_{R}\sim P(R \mid D, \mathcal{P})$ \cite{jorion1997value,brown2018efficient}. 

%We also experiment with using a linear combination of two utility functions $U_\mu$ and $U_{\rm risk}$. By combining these two utilities we can adjust for return and risk similar to the classical Markowitz mean variance formulation for portfolio optimization \cite{markowitz1952portfolio}. Using a combination of expected return and risk allows a spectrum of risk profiles and avoids rewarding policies that are so conservative that they fail to accomplish the task. In related work, \citet{lacotte2019risk} solve for a policy that minimizes its cost subject to a constraint on the conditional value at risk. We consider using a linear combination of risk and policy return as well as the problem of selecting a policy that has highest expected return, subject to a risk threshold.

\section{Experimental Results}

\subsection{Bayesian IRL vs. Bayesian REX}
As noted previously, Bayesian IRL does not scale to high-dimensional tasks due to the requirement of repeatedly solving for an MDP in the inner loop. However, for low-dimensional problems it is still interesting to compare Bayesian IRL with Bayesian REX. We performed a large number of experiments on a variety of randomly generated gridworlds with low-dimensional reward features. We summarize our results here for three different ablations and give full results and implementation details in the Appendix.

\paragraph{Ranked Suboptimal vs. Optimal Demos:} Given a sufficient number of suboptimal ranked demonstrations ($ > 5$), Bayesian REX performs on par and occasionally better than Bayesian IRL when given the same number of optimal demonstrations.

\paragraph{Only Ranked Suboptimal Demos} Bayesian REX always significantly outperforms Bayesian IRL when both algorithms receive suboptimal ranked demonstrations. For fairer comparison, we used a Bayesian IRL algorithm designed to learn from both good and bad demonstrations \cite{cui2018active}. We labeled the top $X$\% ranked demonstrations as good and bottom $X$\% ranked as bad. This improved results for Bayesian IRL, but Bayesian REX still performed significantly better across all $X$.

\paragraph{Only Optimal Demos:} Given a sufficient number of optimal demonstrations ( $> 5$), Bayesian IRL significantly outperforms Bayesian REX. To use Bayesian REX with only optimal demonstrations, we followed prior work \cite{brown2019drex} and  auto-generated pairwise preferences using uniform random rollouts that were labeled as less preferred than the demonstrations. In general, this performed much worse than Bayesian IRL, but for small numbers of demonstrations ( $\leq 5$) Bayesian REX leverages self-supervised rankings to perform nearly as well as full Bayesian IRL.

These results demonstrate that if a very small number of unlabeled near-optimal demonstrations are available, then classical Bayesian IRL is the natural choice for performing reward inference. However, if any of these assumptions are not true, then Bayesian REX is a competitive and often superior alternative for performing Bayesian reward inference even in low-dimensional problems where an MDP solver is tractable. If a highly efficient MDP solver is not available, then Bayesian IRL is infeasible and Bayesian REX is the natural choice for Bayesian reward inference.

\subsection{Visual Imitation Learning via Bayesian REX}
We next tested the imitation learning performance of Bayesian REX for high-dimensional problems where classical Bayesian reward inferernce is infeasible.
%We enforce constraints on the weight vectors by normalizing the output of the weight vector proposal such that $\|w\|_2 = 1$ and use a Gaussian proposal function centered on $w$ with standard deviation $\sigma$. Thus, given the current sample $w_t$, the proposal is defined as $w_{t+1} = \texttt{normalize}(\mathcal{N}(w_t, \sigma))$,
%in which \texttt{normalize} projects the sample back to the surface of the L2-unit ball. %We used models pre-trained from pairwise preferences using T-REX to obtain $\phi(s)$ \cite{browngoo2019trex}.\footnote{pre-trained networks are available at \url{https://github.com/hiwonjoon/ICML2019-TREX/tree/master/atari/learned_models/icml_learned_rewards}} This results in a 65 dimensional feature vector $\phi(s)$. %We initialize the network with these weights and then froze all but the last layer to form $\phi(s)$. 
We pre-trained a 64 dimensional latent state embedding $\phi(s)$ using the self-supervised losses shown in Table~\ref{tab:aux_losses} and the T-REX pairwise preference loss. We found via ablation studies that combining the T-REX loss with the self-supervised losses resulted in better performance than training only with the T-REX loss or only with the self-supervised losses (see Appendix for details). We then used Bayesian REX to generate 200,000 samples from the posterior $P(R \mid D, \mathcal{P})$. To optimize a control policy, we used Proximal Policy Optimization (PPO) \cite{schulman2017proximal} with the MAP and mean reward functions from the posterior  (see Appendix for details).

\begin{table*}
  \caption{Ground-truth average scores when optimizing the mean and MAP rewards found using Bayesian REX. We also compare against the performnace of T-REX \cite{browngoo2019trex} and GAIL \cite{ho2016generative}. Bayesian REX and T-REX are each given 12 demonstrations with ground-truth pairwise preferences. GAIL cannot learn from preferences so it is given 10 demonstrations comparable to the best demonstration given to the other algorithms. The average performance for each IRL algorithm is the average over 30 rollouts.}
  \label{tab:deepbirl_rl}
  \centering
    \vspace{0.1cm}
\begin{tabular}{cccccccc}
\toprule
 & \multicolumn{2}{c}{Ranked Demonstrations} & \multicolumn{1}{c}{Bayesian REX Mean} & \multicolumn{1}{c}{Bayesian REX MAP} & T-REX & GAIL\\
 \midrule
Game &  Best & Avg & Avg (Std) & Avg (Std) & Avg & Avg\\ 
\midrule 
Beam Rider &	1332 &	686.0 &	 5,504.7 (2121.2) &
 \textbf{5,870.3} (1905.1)  & 3,335.7 & 355.5\\
Breakout &	32 &	14.5 &  390.7 (48.8) & \textbf{393.1} (63.7)	 & 221.3 & 0.28\\
Enduro &	84 &	39.8 & 487.7 (89.4)
 & 135.0 (24.8) & \textbf{586.8} & 0.28\\
Seaquest &	600 &	373.3 &  734.7 (41.9)  & 606.0 (37.6) & \textbf{747.3} & 0.0\\ 
Space Invaders &	600 &	332.9 & \textbf{1,118.8} (483.1) & 961.3 (392.3)
 & 1,032.5 & 370.2\\ 
\bottomrule
\end{tabular}
\end{table*}

To test whether Bayesian REX scales to complex imitation learning tasks we selected five Atari games from the Arcade Learning Environment \cite{bellemare2013arcade}. We do not give the RL agent access to the ground-truth reward signal and mask the game scores and number of lives in the demonstrations.
% consisting of stacks of four Atari frames using the standard preprocessing implemented in OpenAIBaselines \cite{baselines}.
Table~\ref{tab:deepbirl_rl} shows the imitation learning performance of Bayesian REX. We also compare against the results reported by \cite{browngoo2019trex} for T-REX, and GAIL \cite{ho2016generative} and use the same 12 suboptimal demonstrations used by \citet{browngoo2019trex} to train Bayesian REX (see Appendix for details). %T-REX uses a sigmoid to normalize rewards before passing them to the RL algorithm; however, we obtained better performance for Bayesian REX by feeding the unnormalized predicted reward $R_\theta(s)$ into PPO for policy optimization.

Table~\ref{tab:deepbirl_rl} shows that Bayesian REX is able to utilize preferences over demonstrations to infer an accurate reward function that enables better-than-demonstrator performance. The average ground-truth return for Bayesian REX surpasses the performance of the best demonstration across all 5 games. In comparison, GAIL seeks to match the demonstrator's state-action distributions which makes imitation learning difficult when demonstrations are suboptimal and noisy.
In addition to providing uncertainty information, Bayesian REX remains competitive with T-REX (which only finds a maximum likelihood estimate of the reward function) and achieves better performance on 3 out of 5 games. %Optimizing the mean reward function as opposed to the MAP reward function  results in superior or comparable performance.% to optimizing the MAP reward function. %This is likely due to the fact that, in theory, optimizing for the mean reward results in a more robust policy \cite{ramachandran2007bayesian}. 
%somehow seaquest did really well with no T-REX, but maybe not for the right reasons...

% \subsection{Show that this works for different eval policies}
% \textbf{TODO}
% Works for different RL algos.
% Works for IRL (T-REX).
% Works for BC.
% Works for GAIL.
% Works for hand-crafted (hacky policies like no-op, act like playing game but don't fire, submerge, release ball, etc)

% \subsection{Black-box policy improvement}
% Maybe show an example of how this could be used to select between pairs of policies a baseline and a potential next policy.

\subsection{High-Confidence Policy Performance Bounds}
Next, we ran an experiment to validate whether the posterior distribution generated by Bayesian REX can be used to solve the HCPE-IL problem described in Section~\ref{sec:hcpe-il}. 
%accurately bound the expected return of different evaluation policies under the unknown reward function $R^*$. 
We evaluated four different evaluation policies, $A \prec B \prec C \prec D$, created by partially training a PPO agent on the ground-truth reward function and checkpointing the policy at various stages of learning. We ran Bayesian REX to generate 200,000 samples from $P(R \mid  D, \mathcal{P})$. To address some of the ill-posedness of IRL, we normalize the weights $w$ such that $\|w\|_2 = 1$. Given a fixed scale for the reward weights, we can compare the relative performance of the different evaluation policies when evaluated over the posterior.% as proposed by \citet{brown2018efficient}. %We see that the modes predicted by Bayesian REX match the ordering of the modes of policies A--D under the true reward function.

\begin{table}[t]
  \caption{Beam Rider policy evaluation bounds compared with ground-truth game scores. Policies A-D correspond to evaluation policies of varying quality obtained by checkpointing an RL agent during training. The No-Op policy seeks to hack the learned reward by always playing the no-op action, resulting in very long trajectories with high mean predicted performance but a very negative 95\%-confidence (0.05-VaR) lower bound on expected return.}
  \label{tab:beamriderPolicyEval}
  \centering
  \vspace{0.1cm}
  \begin{tabular}{ccccc}
    \toprule
    & \multicolumn{2}{c}{Predicted}  & \multicolumn{2}{c}{Ground Truth Avg.}  \\
 Policy & Mean & 0.05-VaR& Score & Length\\
 \midrule
A & 17.1 & 7.9  & 480.6 & 1372.6  \\
B & 22.7 & 11.9  & 703.4 &  1,412.8  \\
C & 45.5 & 24.9  & 1828.5 & 2,389.9  \\
D & 57.6 & 31.5  & 2586.7 & 2,965.0  \\
%Mean & 143.7 & 51.4  & 820.3 & 44 & 8385.7  \\
%MAP & 141.7 & 57.5  & 1756.2 & 264 & 7614.9  \\
No-Op & 102.5 & -1557.1 & 0.0  & 99,994.0  \\
    \bottomrule
  \end{tabular}
\end{table}

\begin{table}[t]
  \caption{Breakout policy evaluation bounds compared with ground-truth game scores. 
  %Policies A-D correspond to evaluation policies of varying quality obtained by checkpointing an RL agent during training. The MAP policies is the result of optimizing a policy via PPO using the MAP reward found via Bayesian REX. 
  Top Half: No-Op never releases the ball, resulting in high mean predicted performance but a low 95\%-confidence bound (0.05-VaR). The MAP policy has even higher risk but also high expected return. Bottom Half: After rerunning MCMC with a ranked trajectory from both the MAP and No-Op policies, the posterior distribution matches the true preferences.}
  \label{tab:breakoutPolicyEval}
  \centering
  \vspace{0.1cm}
  \begin{tabular}{cccccc}
    \toprule
    \midrule
    \multicolumn{6}{c}{Risk profiles given initial preferences} \\
    \midrule
    \midrule
    & \multicolumn{2}{c}{Predicted}  & \multicolumn{2}{c}{Ground Truth Avg.}&  \\
 Policy & Mean & 0.05-VaR& Score & Length \\
 \midrule
A & 1.5 & 0.5  & 1.9 & 202.7  \\
B & 6.3 & 3.7  & 15.8 & 608.4  \\
C & 10.6 & 5.8  & 27.7 & 849.3  \\
D & 13.9 & 6.2  & 41.2 & 1020.8  \\
MAP & 98.2 & -370.2  & 401.0 & 8780.0  \\
No-Op & 41.2 & 1.0  & 0.0 & 7000.0  \\
\midrule
\midrule
\multicolumn{6}{c}{Risk profiles after rankings w.r.t. MAP and No-Op} \\
\midrule
\midrule
A & 0.7 & 0.3  & 1.9 & 202.7  \\
B & 8.7 & 5.5  & 15.8 & 608.4  \\
C & 18.3 & 12.1  & 27.7 & 849.3  \\
D & 26.3 & 17.1  & 41.2 & 1020.8  \\
MAP & 606.8 & 289.1  & 401.0 & 8780.0  \\
No-Op & -5.0 & -13.5  & 0.0 & 7000.0  \\
    \bottomrule
  \end{tabular}
\end{table}

The results for Beam Rider are shown in Table~\ref{tab:beamriderPolicyEval}. We show results for partially trained RL policies A--D. We found that the ground-truth returns for the checkpoints were highly correlated with the mean and 0.05-VaR (5th percentile policy return) returns under the posterior. However, we also noticed that the trajectory length was also highly correlated with the ground-truth reward. If the reward function learned via IRL gives a small positive reward at every time step, then long polices that do the wrong thing may look good under the posterior. To test this hypothesis we used a No-Op policy that seeks to exploit the learned reward function by not taking any actions. This allows the agent to live until the Atari emulator times out after 99,994 steps. 

Table~\ref{tab:beamriderPolicyEval} shows that while the No-Op policy has a high expected return over the chain, looking at the 0.05-VaR shows that the No-Op policy has high risk under the distribution, much lower than evaluation policy A. 
%This finding validates the results by Brown and Niekum \cite{brown2018efficient} that demonstrated the value of using a probabilistic worst-case bound for evaluating the performance of policies when the true reward function is unknown. 
Our results demonstrate that reasoning about probabilistic worst-case performance may be one potential way to detect policies that exhibit so-called reward hacking \cite{amodei2016concrete} or that have overfit to certain features in the demonstrations that are correlated with the intent of the demonstrations, but do not lead to desired behavior, a common problem in imitation learning \cite{ibarz2018reward,de2019causal}.

Table~\ref{tab:breakoutPolicyEval} contains policy evaluation results for the game Breakout. The top half of the table shows the mean return and 95\%-confidence lower bound on the expected return under the reward function posterior for four evaluation policies as well as the MAP policy found via Bayesian IRL and a No-Op policy that never chooses to release the ball. Both the MAP and No-Op policies have high expected returns under the reward function posterior, but also have high risk (low 0.05-VaR). The MAP policy has much higher risk than the No-Op policy, despite good true performance. One likely reason is that, as shown in Table~\ref{tab:deepbirl_rl}, the best demonstrations given to Bayesian REX only achieved a game score of 32. Thus, the MAP policy represents an out of distribution sample and thus has potentially high risk, since Bayesian REX was not trained on policies that hit any of the top layers of bricks. The ranked demonstrations do not give enough evidence to eliminate the possibility that only lower layers of bricks should be hit.%, but they do give strong evidence that missing the ball and quickly losing the game is bad. This results in the No-Op policy have a higher high-confidence performance bound over the posterior.

To test whether active learning can help, we incorporated two active queries: a single rollout from the MAP policy and a single rollout from the No-Op policy and ranked them as better and worse, respectively, than the original set of 12 suboptimal demonstrations. As the bottom of Table~\ref{tab:breakoutPolicyEval} shows, adding two more ranked demonstrations and re-running Bayesian inference, results in a significant change in the risk profiles of the MAP and No-Op policy---the No-Op policy is now correctly predicted to have high risk and low expected returns and the MAP policy now has a much higher 95\%-confidence lower bound on performance.

In the Appendix, we also compare the performance of Bayesian REX to alternative methods such as using MC dropout \cite{gal2016dropout} or using an ensemble \cite{lakshminarayanan2017simple} to estimate uncertainty and compute high-confidence performance bounds.

%TODO: Add Beamrider and Space invaders human demo evals

\subsection{Human Demonstrations}
To investigate whether Bayesian REX is able to correctly rank human demonstrations, we used Bayesian REX to calculate high-confidence performance bounds for a variety of human demonstrations (see the Appendix for full details and additional results).

We generated four human demonstrations for Beam Rider: (1) \textit{good}, a good demonstration that plays the game well, (2) \textit{bad}, a bad demonstration that seeks to play the game but does a poor job, (3) \textit{pessimal}, a demonstration that does not shoot enemies and seeks enemy bullets, and (4) \textit{adversarial} a demonstration that pretends to play the game by moving and shooting but tries to avoid actually shooting enemies. The resulting high-confidence policy evaluations are shown in Table~\ref{tab:beamRiderHumanDemoPolicyEval}. The high-confidence bounds and average performance over the posterior correctly rank the behaviors. This provides evidence that the learned linear reward correctly rewards actually destroying aliens and avoiding getting shot, rather than just flying around and shooting.

\begin{table}
  \caption{Beam Rider human demonstrations.}
  \label{tab:beamRiderHumanDemoPolicyEval}
  \centering
  \vspace{0.1cm}
  \begin{tabular}{ccccc}
    \toprule
    & \multicolumn{2}{c}{Predicted}  & \multicolumn{2}{c}{Ground Truth}  \\
 Policy & Mean & 0.05-VaR& Avg. & Length \\
 \midrule
good & 12.4 & 5.8 & 1092 & 1000.0 \\
bad & 10.7 & 4.5 & 396 & 1000.0 \\
pessimal & 6.6 & 0.8 & 0 & 1000.0 \\
adversarial & 8.4 & 2.4 & 176 & 1000.0 \\
\bottomrule
  \end{tabular}
\end{table}

Next we demonstrated four different behaviors when playing Enduro: (1) \textit{good} a demonstration that seeks to play the game well, (2) \textit{periodic} a demonstration that alternates between speeding up and passing cars and then slowing down and being passed, (3) \textit{neutral} a demonstration that stays right next to the last car in the race and doesn't try to pass or get passed, and (4) \textit{ram} a demonstration that tries to ram into as many cars while going fast. Table~\ref{tab:enduroHumanDemoPolicyEval} shows that Bayesian REX is able to accurately predict the performance and risk of each of these demonstrations and gives the highest (lowest 0.05-VaR) risk to the \textit{ram} demonstration and the least risk to the \textit{good} demonstration.

\begin{table}
  \caption{Enduro evaluation of a variety of human demonstrations.}
  \label{tab:enduroHumanDemoPolicyEval}
  \centering
  \vspace{0.1cm}
  \begin{tabular}{ccccc}
    \toprule
    & \multicolumn{2}{c}{Predicted}  & \multicolumn{2}{c}{Ground Truth}  \\
 Policy & Mean & 0.05-VaR& Avg. & Length \\
 \midrule
good & 246.7 & -113.2 & 177 & 3325.0 \\
periodic & 230.0 & -130.4 &  44 & 3325.0 \\
neutral & 190.8 & -160.6 & 0 & 3325.0 \\
ram & 148.4 & -214.3 & 0 & 3325.0 \\
\bottomrule
  \end{tabular}
\end{table}

%\subsection{Additional Experiments}
%We also tested Bayesian REX on a wide variety of human demonstrations and found that Bayesian REX is able to robustly rank evaluation policies relative to each other, even when some of the demonstrations are deceptive, e.g., moving and firing but not destroying enemies. We also conducted a more rigorous exploration of the risk-return trade-off for safe policy selection by varying the risk tolerance $\delta$ from 0 (maximally risk averse) to 1 (maximally risk tolerant) and then performing high-confidence policy selection based on the desired risk tolerance threshold. Due to space constraints we have included these additional experimental results in the Appendix.

\section{Conclusion}
Bayesian reasoning is a powerful tool when dealing with uncertainty and risk; however, existing Bayesian reward learning algorithms often require solving an MDP in the inner loop, rendering them intractable for complex problems in which solving an MDP may take several hours or even days. In this paper we propose a novel deep learning algorithm, Bayesian Reward Extrapolation (Bayesian REX), that leverages preference labels over demonstrations to make Bayesian reward inference tractable for high-dimensional visual imitation learning tasks. Bayesian REX can sample tens of thousands of reward functions from the posterior in a matter of minutes using a consumer laptop. We tested our approach on five Atari imitation learning tasks and showed that Bayesian REX achieves state-of-the-art performance in 3 out of 5 games. Furthermore, Bayesian REX enables efficient high-confidence performance bounds for arbitrary evaluation policies. We demonstrated that these high-confidence bounds allow an agent to accurately rank different evaluation policies and provide a potential way to detect reward hacking and value misalignment. 

We note that our proposed safety bounds are only safe with respect to the assumptions that we make: good feature pre-training, rapid MCMC mixing, and accurate preferences over demonstrations. Future work includes using exploratory trajectories for better pre-training of the latent feature embeddings, developing methods to determine when a relevant feature is missing from the learned latent space, and using high-confidence performance bounds to perform safe policy optimization in the imitation learning setting. %Given a starting policy $\pi$ we want to optimize a policy such that it maximizes some safety threshold. %Brown et al. \cite{brown2018efficient} demonstrate an example of this on a simple grid world in which the goal is to find a policy that minimizes value-at-risk. 
\section*{Acknowledgements}
This work has taken place in the Personal Autonomous Robotics Lab (PeARL) at The University of Texas at Austin. PeARL research is supported in part by the NSF (IIS-1724157, IIS-1638107,IIS-1617639, IIS-1749204) and ONR(N00014-18-2243, N00014-17-1-2143). This research was also sponsored by the Army Research Office and was accomplished under Cooperative Agreement Number W911NF-19-2-0333. The views and conclusions contained in this document are those of the authors and should not be interpreted as representing the official policies, either expressed or implied, of the Army Research Office or the U.S. Government. The U.S. Government is authorized to reproduce and distribute reprints for Government purposes notwithstanding any copyright notation herein.

% \textbf{Do not} include acknowledgements in the initial version of
% the paper submitted for blind review.

% If a paper is accepted, the final camera-ready version can (and
% probably should) include acknowledgements. In this case, please
% place such acknowledgements in an unnumbered section at the
% end of the paper. Typically, this will include thanks to reviewers
% who gave useful comments, to colleagues who contributed to the ideas,
% and to funding agencies and corporate sponsors that provided financial

% In the unusual situation in which you want a paper to appear in the

\bibliography{safeirl}
\bibliographystyle{icml2020}

 \appendix
  
 \section{Source Code and Videos}
 See the project webpage \url{https://sites.google.com/view/bayesianrex/}. Code repo for the Atari experiments is available at \url{https://github.com/dsbrown1331/bayesianrex}. Code for the gridworld experiments can be found at \url{https://github.com/dsbrown1331/brex_gridworld_cpp}
 
 \section{MCMC Details}

We represent $R_\theta$ as a linear combination of pre-trained features:
\begin{equation}
R_\theta (\tau) = \sum_{s \in \tau} w^T\phi(s) = w^T\sum_{s \in \tau} \phi(s) = w^T \Phi_{\tau}.
\end{equation}
We pre-compute and cache $\Phi_{\tau_i} = \sum_{s \in \tau_i} \phi(s)$ for $i = 1,\ldots,m$ and the likelihood becomes
\begin{equation}\label{eqn:lincombo_Boltzman_app}
P(\mathcal{P}, D  \mid R_\theta) = \prod_{(i,j) \in \mathcal{P}} \frac{e^{\beta w^T \Phi_{\tau_j}}}{e^{\beta w^T \Phi_{\tau_j}} + e^{\beta w^T \Phi_{\tau_i}}}.
\end{equation}

We use $\beta=1$ and enforce constraints on the weight vectors by normalizing the output of the weight vector proposal such that $\|w\|_2 = 1$ and use a Gaussian proposal function centered on $w$ with standard deviation $\sigma$. Thus, given the current sample $w_t$, the proposal is defined as $w_{t+1} = \texttt{normalize}(\mathcal{N}(w_t, \sigma))$,
in which \texttt{normalize} divides by the L2 norm of the sample to project back to the surface of the L2-unit ball.

For all experiments, except Seaquest, we used a default step size of 0.005. For Seaquest increased the step size to 0.05. We run 200,000 steps of MCMC and use a burn-in of 5000 and skip every 20th sample to reduce auto-correlation. We initialize the MCMC chain with a randomly chosen vector on the L2-unit ball. Because the inverse reinforcement learning is ill-posed there are an infinite number of reward functions that could match any set of demonstrations. Prior work by \citet{finn2016guided} demonstrates that strong regularization is needed when learning cost functions via deep neural networks. To ensure that the rewards learned allow good policy optimization when fed into an RL algorithm we used a non-negative return prior on the return of the lowest ranked demonstration. The prior takes the following form:
\begin{equation}
\log P(w) = \begin{cases}
0 & \text{if } e^{\beta w^T \Phi_{\tau_1}} < 0 \\
-\infty & \text{otherwise}
\end{cases}
\end{equation}
This forces MCMC to not only find reward function weights that match the rankings, but to also find weights such that the return of the worse demonstration is non-negative. If the return of the worse demonstration was negative during proposal generation, then we assigned it a prior probability of $-\infty$. Because the ranking likelihood is invariant to affine transformations of the rewards, this prior simply shifts the range of learned returns and does not affect the log likelihood ratios.

\begin{algorithm}[t]
\caption{Bayesian REX: Bayesian Reward Extrapolation}
\label{alg:DeeP-BIRL}
\begin{algorithmic}[1]
\STATE \textbf{Input:} demonstrations $D$, pairwise preferences $\mathcal{P}$, MCMC proposal width $\sigma$, number of proposals to generate $N$, deep network architecture $R_\theta$, and prior $P(w)$.
\STATE pre-train $R_\theta$ using auxiliary tasks (see Section~5.2).
\STATE Freeze all but last layer, $w$, of $R_\theta$. 
\STATE $\phi(s) \vcentcolon=$ activations of the penultimate layer of $R_\theta$.
\STATE Precompute and cache $\Phi_{\tau} = \sum_{s \in \tau} \phi(s)$ for all $\tau \in D$.
\STATE Initialize $w$ randomly.
\STATE Chain[0] $\gets  w$
\STATE Compute $P(\mathcal{P}, D | w) P(w)$ using Equation~(\ref{eqn:lincombo_Boltzman_app})
\FOR{$i \gets 1$ to $N$}
	\STATE $\tilde{w} \gets \texttt{normalize}(\mathcal{N}(w_t, \sigma))$
	\STATE Compute $P(\mathcal{P}, D | \tilde{w}) P(\tilde{w})$ using Equation~(\ref{eqn:lincombo_Boltzman_app})
		\STATE	$u \gets \texttt{Uniform} (0,1)$
	\IF {$\displaystyle u < \frac{P(\mathcal{P}, D | \tilde{w}) P(\tilde{w})}{P(\mathcal{P}, D | w) P(w)}$} 
    	\STATE Chain[i] $\gets \tilde{w}$
    	\vspace{.5mm}
    	\STATE $ w \gets \tilde{w}$
	\ELSE
		\STATE Chain[i] $\gets w$
	\ENDIF
\ENDFOR
\STATE \textbf{return} Chain
\end{algorithmic}
\end{algorithm}

\section{Bayesian IRL vs. Bayesian REX}
Bayesian IRL does not scale to high-dimensional tasks due to the requirement of repeatedly solving for an MDP in the inner loop. In this section we focus on low-dimensional problems where it is tractable to repeatedly solve an MDP. We compare the performance of Bayesian IRL with Bayesian REX when performing reward inference. Because both algorithms make very different assumptions, we compare their performance across three different tasks. The first task attempts to give both algorithms the demonstrations they were designed for. The second evaluation focuses on the case where all demonstrations are optimal and is designed to put Bayesian IRL at a disadvantage. The third evaluation focuses on the case where all demonstrations are optimal and is designed to put Bayesian REX at a disadvantage. Note that we focus on sample efficiency rather than computational efficiency as Bayesian IRL is significantly slower than Bayesian REX as it requires repeatedly solving an MDP, whereas Bayesian REX requires no access to an MDP during reward inference.

All experiments were performed using 6x6 gridworlds with 4 binary features placed randomly in the environment. The ground-truth reward functions are sampled uniformly from the L1-ball \cite{brown2018efficient}. The agent can move in the four cardinal directions and stays in place if it attempts to move off the grid. Transitions are deterministic, $\gamma = 0.9$, and there are no terminal states. We perform evaluations over 100 random gridworlds for varying numbers of demonstrations. Each demonstration is truncated to a horizon of 20. We use $\beta = 50$ for both Bayesian IRL and Bayesian REX and we remove duplicates from demonstrations. After performing MCMC we used a 10\% burn-in period for both algorithms and only used every 5th sample after the burn-in when computing the mean reward under the posterior. We then optimized a policy under the mean reward from the Bayesian IRL posterior and under the mean reward from the Bayesian REX posterior. We then compare the average policy loss for each algorithm when compared with optimal performance under the ground-truth reward function. 

Code for the gridworld experiments can be found at \url{https://github.com/dsbrown1331/brex_gridworld_cpp}.

\subsection{Ranked Suboptimal vs. Optimal Demonstrations}
We first compare Bayesian IRL when it is given varying numbers of optimal demonstrations with Bayesian REX when it receives the same number of suboptimal demonstrations. We give each algorithm the demonstrations best suited for its assumptions while keeping the number of demonstrations equal and using the same starting states for each algorithm. To generate suboptimal demonstrations we simply use random rollouts and then rank them according to the ground-truth reward function.

Table~\ref{tab:subopt_vs_opt_app} shows that, given a sufficient number of suboptimal ranked demonstrations ($ > 5$), Bayesian REX performs on par or slightly better than Bayesian IRL when given the same number of optimal demonstrations starting from the same states as the suboptimal demonstrations. This result shows that not only is Bayesian REX much more computationally efficient, but it also has sample efficiency comparable to Bayesian IRL as long as there are a sufficient number of ranked demonstrations. Note that 2 ranked demonstrations induces only a single constraint on the reward function so it is not surprising that it performs much worse than running full Bayesian IRL with all the counterfactuals afforded by running an MDP solver in the inner-loop.

\begin{table}
  \caption{Ranked Suboptimal vs. Optimal Demos: Average policy loss over 100 random 6x6 grid worlds with 4 binary features for varying numbers of demonstrations.}
  \label{tab:subopt_vs_opt_app}
  \centering
    \vspace{0.1cm}
\begin{tabular}{cccccc}
\toprule
& \multicolumn{5}{c}{Number of demonstrations} \\
    & 2& 5& 10& 20& 30\\
    \midrule
B-IRL & \textbf{0.044}& \textbf{0.033}& 0.020& 0.009& \textbf{0.006}\\
B-REX & 1.779& 0.421& \textbf{0.019}& \textbf{0.006}& \textbf{0.006}\\
\bottomrule
\end{tabular}
\end{table}

\subsection{Only Ranked Suboptimal Demonstrations} 
For the next experiment we consider what happens when Bayesian IRL recieves suboptimal ranked demonstrations. Table~\ref{tab:subopt_birl_brex_app} shows that B-REX always significantly outperforms Bayesian IRL when both algorithms receive suboptimal ranked demonstrations. To achieve a fairer comparison, we also compared Bayesian REX with a Bayesian IRL algorithm designed to learn from both good and bad demonstrations \cite{cui2018active}. We labeled the top $x$\% ranked demonstrations as good and bottom $x$\% ranked as bad. Table~\ref{tab:subopt_yuchen_brex_app} shows that leveraging the ranking significantly improves the performance of Bayesian IRL, but Bayesian REX still performed significantly better across all $x$.

\begin{table}
  \caption{Ranked Suboptimal Demos: Average policy loss for Bayesian IRL versus Bayesian REX over 100 random 6x6 grid worlds with 4 binary features for varying numbers of demonstrations}
  \label{tab:subopt_birl_brex_app}
  \centering
    \vspace{0.1cm}
\begin{tabular}{cccccc}
\toprule
& \multicolumn{5}{c}{Number of demonstrations} \\
& 2& 5& 10& 20& 30\\
\midrule
B-IRL & 3.512& 3.319& 2.791& 3.078& 3.158\\
B-REX & \textbf{1.796}& \textbf{0.393}& \textbf{0.026}& \textbf{0.006}& \textbf{0.006}\\
\bottomrule
\end{tabular}
\end{table}

\begin{table}
  \caption{Ranked Suboptimal Demos: Average policy loss for Bayesian REX and  Bayesian IRL using the method proposed by \cite{cui2018active}* which makes use of good and bad demonstrations. We used the top $x$\% of the ranked demos as good and bottom $x$\% as bad. Results are averaged over 100 random 6x6 grid worlds with 4 binary features.}
  \label{tab:subopt_yuchen_brex_app}
  \centering
    \vspace{0.1cm}
\begin{tabular}{ccccc}
\toprule
& \multicolumn{4}{c}{Top/bottom percent of 20 ranked demos} \\
& x=5\%& x=10\%& x=25\%& x=50\%\\
\midrule
B-IRL($x$)* & 1.120& 0.843& 1.124& 2.111\\
B-REX & \multicolumn{4}{c}{\textbf{0.006}}\\
\bottomrule
\end{tabular}
\end{table}

\subsection{Only Optimal Demonstrations} 
Finally, we compared Bayesian REX with Bayesian IRL when both algorithms are given optimal demonstrations. As an attempt to use Bayesian REX with only optimal demonstrations, we followed prior work \cite{brown2019drex} and auto-generated pairwise preferences using uniform random rollouts that are labeled as less preferred than the demonstrations. Table~\ref{tab:opt_birl_brex_app} shows that Bayesian IRL outperforms Bayesian REX. This demonstrates the value of giving a variety of ranked trajectories to Bayesian REX. For small numbers of optimal demonstrations ( $\leq 5$) we found that Bayesian REX leveraged the self-supervised rankings to only perform slightly worse than full Bayesian IRL. This result is encouraging since it is possible that a more sophisticated method for auto-generating suboptimal demonstrations and rankings could be used to further improve the performance of Bayesian REX even when demonstrations are not ranked \cite{brown2019drex}.

\begin{table}
  \caption{Optimal Demos: Average policy loss for Bayesian IRL versus Bayesian REX over 100 random 6x6 grid worlds with 4 binary features for varying numbers of demonstrations. Bayesian REX uses random rollouts to automatically create preferences (optimal demos are preferred to random rollouts).}
  \label{tab:opt_birl_brex_app}
  \centering
    \vspace{0.1cm}
\begin{tabular}{cccccc}
\toprule
& \multicolumn{5}{c}{Number of demonstrations} \\
& 2& 5& 10& 20& 30\\
\midrule
B-IRL & \textbf{0.045}& \textbf{0.034}& \textbf{0.018}& \textbf{0.009}& \textbf{0.006}\\
B-REX & 0.051& 0.045& 0.040& 0.034& 0.034\\
\bottomrule
\end{tabular}
\end{table}

\subsection{Summary}
The results above demonstrate that if a very small number of unlabeled near-optimal demonstrations are available, then classical Bayesian IRL is the natural choice for performing reward inference. However, if any of these assumptions are not true, then Bayesian REX is a competitive and often superior alternative for performing Bayesian reward inference. Also implicit in the above results is the assumption that a highly tractable MDP solver is available for performing Bayesian IRL. If this is not the case, then Bayesian IRL is infeasible and Bayesian REX is the natural choice for Bayesian reward inference.

\section{Pre-training Latent Reward Features}

We experimented with several pretraining methods. One method is to train $R_\theta$ using the pairwise ranking loss
\begin{equation}\label{eqn:pairwiserank_appendix}
P(D, \mathcal{P}   \mid R_\theta) = \prod_{(i,j) \in \mathcal{P}} \frac{e^{\beta R_\theta(\tau_j)}}{e^{\beta R_\theta(\tau_i)} + e^{\beta R_\theta(\tau_j)}},
\end{equation}
and then freeze all but the last layer of weights; however, the learned embedding may overfit to the limited number of preferences over demonstrations and fail to capture features relevant to the ground-truth reward function. Thus, we supplement the pairwise ranking objective with auxiliary objectives that can be optimized in a self-supervised fashion using data from the demonstrations. 

We use the following self-supervised tasks to pre-train $R_\theta$: (1) Learn an inverse dynamics model that uses embeddings $\phi(s_t)$ and $\phi(s_{t+1})$ to predict the corresponding action $a_t$ \cite{torabi2018behavioral,hanna2017grounded}, (2) Learn a forward dynamics model that predicts $s_{t+1}$ from $\phi(s_t)$ and $a_t$ \cite{oh2015action,thananjeyan2019safety}, (3) Learn an embedding $\phi(s)$ that predicts the temporal distance between two randomly chosen states from the same demonstration \cite{imitationyoutube}, and (4) Train a variational pixel-to-pixel autoencoder in which $\phi(s)$ is the learned latent encoding \cite{makhzani2017pixelgan,doersch2016tutorial}. Table~\ref{tab:aux_losses_app} summarizes the auxiliary tasks used to train $\phi(s)$. 

There are many possibilities for pre-training $\phi(s)$; however, we found that each objective described above encourages the embedding to encode different features. For example, an accurate inverse dynamics model can be learned by only attending to the movement of the agent. Learning forward dynamics supplements this by requiring $\phi(s)$ to encode information about short-term changes to the environment. Learning to predict the temporal distance between states in a trajectory forces $\phi(s)$ to encode long-term progress. Finally, the autoencoder loss acts as a regularizer to the other losses as it seeks to embed all aspects of the state.

\begin{table}
  \caption{Self-supervised learning objectives used to pre-train $\phi(s)$.}
  \label{tab:aux_losses_app}
  \centering
    \vspace{0.1cm}
\begin{tabular}{ll}
\toprule
%Pairwise Ranking &  $f(\phi(\tau_i), \phi(\tau_j)) \rightarrow \mathbf{1}_{\tau_i \succ \tau_j} $ \\
 Inverse Dynamics & $f_{\rm ID}(\phi(s_t), \phi(s_{t+1})) \rightarrow a_t$ \\
 Forward Dynamics & $f_{\rm FD}(\phi(s_t), a_t) \rightarrow s_{t+1}$ \\
 Temporal Distance & $f_{\rm TD}(\phi(s_t), \phi(s_{t+x}) \rightarrow x$ \\
 Variational Autoencoder & $f_{A}(\phi(s_t)) \rightarrow s_t$ \\
\bottomrule
\end{tabular}
\end{table}

\begin{figure*}[t]
% \begin{subfigure}{.24\textwidth}
% \centering
\includegraphics[width=\linewidth]{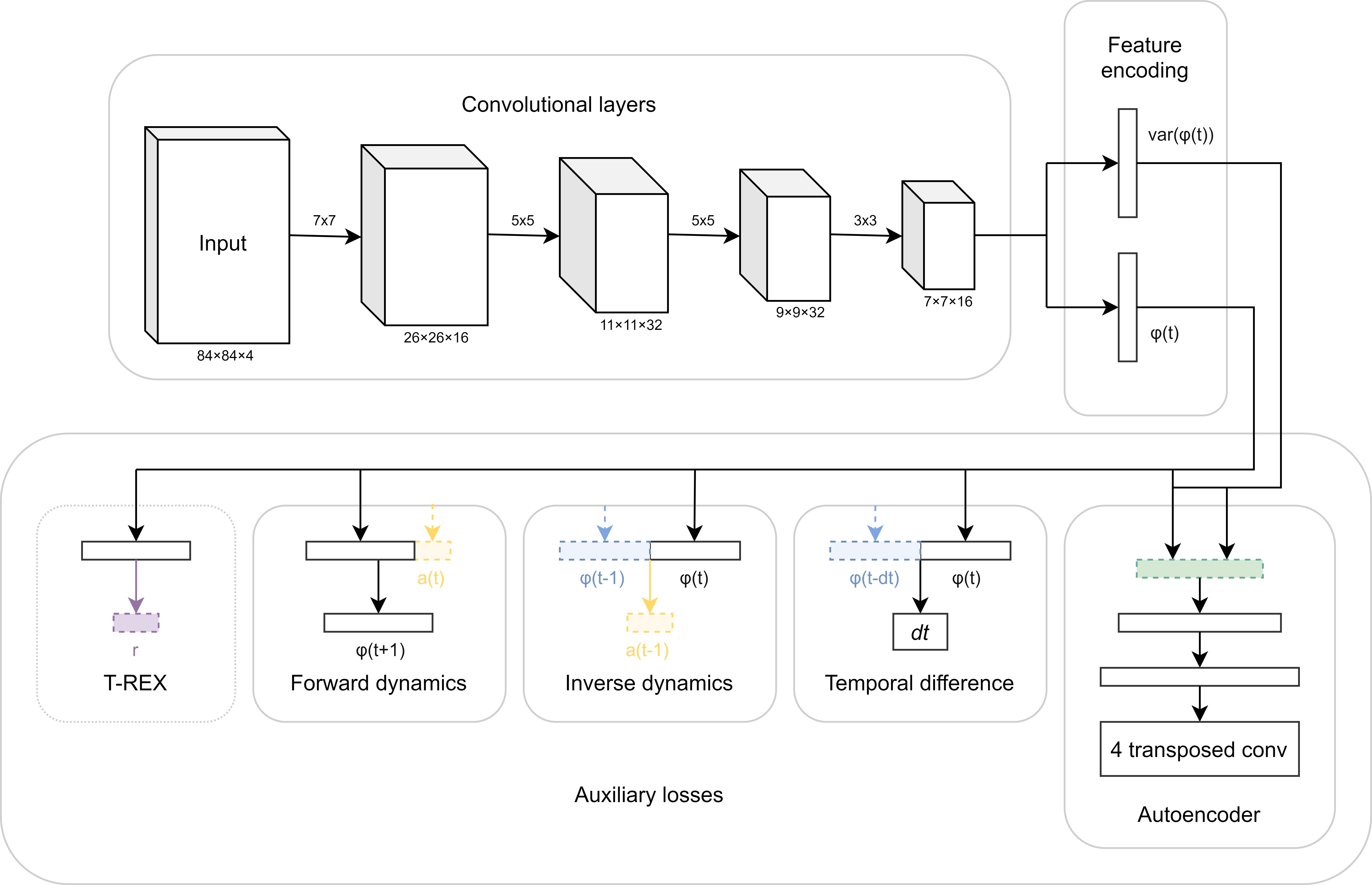}
 \caption{Diagram of the network architecture used when training feature encoding $\phi(s)$ with self-supervised and T-REX losses. Yellow denotes actions, blue denotes feature encodings sampled from elsewhere in a demonstration trajectory, and green denotes random samples for the variational autoencoder.}
%  \label{fig:network_diagram_picture}
\label{fig:network_diagram}
\end{figure*}

In the Atari domain, input to the network is given visually as grayscale frames resized to $84 \times 84$. To provide temporal information, four sequential frames are stacked one on top of another to create a \textit{framestack} which provides a brief snapshot of activity. The network architecture takes a framestack, applies four convolutional layers following a similar architecture to \citet{christiano2017deep} and \citet{browngoo2019trex}, with leaky ReLU units as non-linearities following each convolution layer. The convolutions follow the following structure:
\begin{center}
  %\caption{Structure of four initial convolutions}
  \label{tab:initial_convolutions}
    \begin{tabular}{ c | c c c }
     \# & Filter size & Image size & Stride \\ \hline
     Input & - & $84 \times 84 \times 4$ & - \\  
     1 & $7x7$ & $26 \times 26 \times 16$ & 3 \\
     2 & $5x5$ & $11 \times 11 \times 32$ & 2 \\
     3 & $5x5$ & $9 \times 9 \times 32$ & 1\\
     4 & $3x3$ & $7 \times 7 \times 16$ & 1
    \end{tabular}
\end{center}
The convolved image is then flattened. Two sequential fully connected layers, with leaky ReLU applied to the first layer, transform the flattened image into the encoding, $\phi(s)$ where $s$ is the initial framestack. The width of these layers depends on the size of the feature encoding chosen. In our experiments with a latent dimension of 64, the first layer transforms from size 784 to 128 and the second from 128 to 64.
 
 See Figure~\ref{fig:network_diagram} for a complete diagram of this process.
 
 Architectural information for each auxiliary task is given below.
 \begin{enumerate}
     \item The variational autoencoder (VAE) tries to reconstruct the original framestack from the feature encoding using transposed convolutions. Mirroring the structure of the initial convolutions, two fully connected layers precede four transposed convolution layers. These first two layers transform the 64-dimensional feature encoding from 64 to 128, and from 128 to 1568. The following four layers' structures are summarized below:
     \begin{center}
        \label{tab:transposed_convolutions}
        \begin{tabular}{ c | c c c }
         \# & Filter size & Image size & Stride \\ \hline
         Input & - & $28 \times 28 \times 2$ & - \\  
         1 & $3x3$ & $30 \times 30 \times 4$ & 1 \\
         2 & $6x6$ & $35 \times 35 \times 16$ & 1 \\
         3 & $7x7$ & $75 \times 75 \times 16$ & 2\\
         4 & $10x10$ & $84 \times 84 \times 4$ & 1
        \end{tabular}
    \end{center}
    A cross-entropy loss is applied between the reconstructed image and the original, as well as a term added to penalize the KL divergence of the distribution from the unit normal.
     
     \item A temporal difference estimator, which takes two random feature encodings from the same demonstration and predicts the number of timesteps in between. It is a single fully-connected layer, transforming the concatenated feature encodings into a scalar time difference. A mean-squared error loss is applied between the real difference and predicted.
     \item An inverse dynamics model, which takes two sequential feature encodings and predicts the action taken in between. It is again a single fully-connected layer, trained as a classification problem with a binary cross-entropy loss over the discrete action set.
     \item A forward dynamics model, which takes a concatenated feature encoding and action and predicts the next feature encoding with a single fully-connected layer. This is repeated 5 times, which increases the difference between the initial and final encoding. It is trained using a mean-squared error between the predicted and real feature encoding.
     \item A T-REX loss, which samples feature encodings from two different demonstrations and tries to predict which one of them has preference over the other. This is done with a single fully-connected layer that transforms an encoding into scalar reward, and is then trained as a classification problem with a binary cross-entropy loss. A 1 is assigned to the demonstration sample with higher preference and a 0 to the demonstration sample with lower preference.
 \end{enumerate}
 In order to encourage a feature encoding that has information easily interpretable via linear combinations, the temporal difference, T-REX, inverse dynamics, and forward dynamics tasks consist of only a single layer atop the feature encoding space rather than multiple layers.

 To compute the final loss on which to do the backwards pass, all of the losses described above are summed with weights determined empirically to balance out their values. 
 
 \subsection{Training specifics}
 We used an NVIDIA TITAN V GPU for training the embedding. We used the same 12 demonstrations used for MCMC to train the self-supervised and ranking losses described above. We sample 60,000 trajectory snippets pairs from the demonstration pool, where each snippet is between 50 and 100 timesteps long. We use a learning rate of 0.001 and a weight decay of 0.001. We make a single pass through all of the training data using batch size of 1 resulting in 60,000 updates using the Adam \cite{kingma2014adam} optimizer. For Enduro prior work \cite{browngoo2019trex} showed that full trajectories resulted in better performance than subsampling trajectories. Thus, for Enduro we subsample 10,000 pairs of entire trajectories by randomly selecting a starting time between 0 and 5 steps after the initial state and then skipping every t frames where t is chosen uniformly from the range $[3,7)$ and train with two passes through the training data. When performing subsampling for either snippets or full trajectories, we subsample pairs of trajectories such that one is from a worse ranked demonstration and one is from a better ranked demonstration following the procedure outlined in \cite{browngoo2019trex}.

\section{Visualizations of Learned Features}
Viewable \href{https://www.youtube.com/watch?v=DMf8kNH9nVg}{here}\footnote{\url{https://www.youtube.com/watch?v=DMf8kNH9nVg}} is a video containing an Enduro demonstration trajectory, its decoding with respect to the pre-trained autoencoder, and a plot of the dimensions in the latent encoding over time. Observe how changes in the demonstration, such as turning right or left or a shift, correspond to changes in the plots of the feature embedding. We noticed that certain features increase when the agent passes other cars while other features decrease when the agent gets passed by other cars. This is evidence that the pretraining has learned features that are relevant to the ground truth reward which gives +1 every time the agent passes a car and -1 every time the agent gets passed.

Viewable \href{https://www.youtube.com/watch?v=2uN5uD17H6M}{here}\footnote{\url{https://www.youtube.com/watch?v=2uN5uD17H6M}} is a similar visualization of the latent space for Space Invaders. Notice how it tends to focus on the movement of enemy ships, useful for game progress in things such as the temporal difference loss, but seems to ignore the player's ship despite its utility in inverse dynamics loss. Likely the information exists in the encoding but is not included in the output of the autoencoder.

Viewable \href{https://www.youtube.com/watch?v=8zgbD1fZOH8}{here}\footnote{\url{https://www.youtube.com/watch?v=8zgbD1fZOH8}} is visualization of the latent space for Breakout. Observe that breaking a brick often results in a small spike in the latent encoding. Many dimensions, like the dark green curve which begins at the lowest value, seem to invert as game progress continues on, thus acting as a measure of how much time has passed.
 
\section{Imitation Learning Ablations for Pre-training $\phi(s)$}

Table~\ref{tab:deepbirl_rl_trex} shows the results of pre-training reward features only using different losses. We experimented with using only the T-REX Ranking loss \cite{browngoo2019trex}, only the self-supervised losses shown in Table 1 of the main paper, and using both the T-REX ranking loss plus the self-supervised loss function. We found that performance varried over the different pre-training schemes, but that using Ranking + Self-Supervised achieved high performance across all games, clearly outperforming only using self-supervised losses and achieving superior performance to only using the ranking loss on 3 out of 5 games.

\begin{table*}
  \caption{Comparison of different reward feature pre-training schemes. Ground-truth average returns for several Atari games when optimizing the mean and MAP rewards found using Bayesian REX. Each algorithm is given the same 12 demonstrations with ground-truth pairwise preferences. The average performance for each IRL algorithm is the average over 30 rollouts.}
  \label{tab:deepbirl_rl_trex}
  \centering
    \vspace{0.1cm}
\begin{tabular}{ccccccc}
\toprule
 & \multicolumn{2}{c}{Ranking Loss} & \multicolumn{2}{c}{Self-Supervised} & \multicolumn{2}{c}{Ranking + Self-Supervised}\\
 \midrule
Game &  Mean & MAP & Mean & MAP & Mean & MAP\\ 
\midrule 
Beam Rider & 3816.7  & 4275.7  & 180.4  & 143.7  & \textbf{5870.3}  & 5504.7 \\ 
Breakout & 389.9  & \textbf{409.5}  & 360.1  & 367.4  & 393.1  & 390.7 \\ 
Enduro & 472.7  & 479.3  & 0.0  & 0.0  & 135.0  & \textbf{487.7} \\ 
Seaquest & 675.3  & 670.7  & 674.0  & 683.3  & 606.0  & \textbf{734.7} \\ 
Space Invaders & \textbf{1482.0}  & 1395.5  & 391.2  & 396.2  & 961.3  & 1118.8 \\
\bottomrule
\end{tabular}
\end{table*}

\section{Suboptimal Demonstration Details}
We used the same suboptimal demonstrations used by \citet{browngoo2019trex} for comparison. These demonstrations were obtained by running PPO on the ground truth reward and checkpointing every 50 updates using OpenAI Baselines \cite{baselines}. \citet{browngoo2019trex} make the checkpoint files available, so to generate the demonstration data we used their saved checkpoints and followed the instructions in their released code to generate the data for our algorithm\footnote{Code from \cite{browngoo2019trex} is available here \url{https://github.com/hiwonjoon/ICML2019-TREX}}. We gave Bayesian REX these demonstrations as well as ground-truth rankings using the game score; however, other than the rankings, Bayesian REX has no access to the true reward samples. Following the recommendations of \citet{browngoo2019trex}, we mask the game score and other parts of the game that are directly indicative of the game score such as the number of enemy ships left, the number of lives left, the level number, etc. See \cite{browngoo2019trex} for full details.

\section{Reinforcement Learning Details}
We used the OpenAI Baselines implementation of Proximal Policy Optimization (PPO) \cite{schulman2017proximal,baselines}. We used the default hyperparameters for all games and all experiments. We run RL for 50 million frames and then take the final checkpoint to perform evaluations.  We adapted the OpenAI Baselines code so even though the RL agent receives a standard preprocessed observation, it only receives samples of the reward learned via Bayesian REX, rather than the ground-truth reward. T-REX \cite{browngoo2019trex} uses a sigmoid to normalize rewards before passing them to the RL algorithm; however, we obtained better performance for Bayesian REX by feeding the unnormalized predicted reward $R_\theta(s)$ into PPO for policy optimization. We follow the OpenAI baselines default preprocessing for the framestacks that are fed into the RL algorithm as observations. We also apply the default OpenAI baselines wrappers the environments. We run PPO with 9 workers on an NVIDIA TITAN V GPU.

\section{High-Confidence Policy Performance Bounds}
In this section we describe the details of the policy performance bounds.

\subsection{Policy Evaluation Details}
We estimated $\Phi_{\pi_{\rm eval}}$ using $C$ Monte Carlo rollouts for each evaluation policy. Thus, after generating $C$ rollouts, $\tau_1,\ldots,\tau_C$ from $\pi_{\rm eval}$ the feature expectations are computed as
\begin{equation}
\Phi_{\pi_{\rm eval}}  = \frac{1}{C}\left[ \sum_{i=1}^C \sum_{s \in \tau_i} \phi(s)  \right].
\end{equation}
We used $C=100$ for all experiments.

\subsection{Evaluation Policies}
We evaluated several different evaluation policies. To see if the learned reward function posterior can interpolate and extrapolate we created four different evaluation policies: A, B, C, and D. These policies were created by running RL via PPO on the ground truth reward for the different Atari games. We then checkpointed the policy and selected checkpoints that would result in different levels of performance. For all games except for Enduro these checkpoints correspond to 25, 325, 800, and 1450 update steps using OpenAI baselines. For Enduro, PPO performance was stuck at 0 return until much later in learning. To ensure diversity in the evaluation policies, we chose to use evaluation policies corresponding to 3125, 3425, 3900, and 4875 steps. We also evaluated each game with a No-Op policy. These policies are often adversarial for some games, such as Seaquest, Breakout, and Beam Rider, since they allow the agent to live for a very long time without actually playing the game---a potential way to hack the learned reward since most learned rewards for Atari will incentivize longer gameplay.

The results for Beam Rider and Breakout are shown in the main paper. For completeness, we have included the high-confidence policy evaluation results for the other games here in the Appendix. Table~\ref{tab:enduroPolicyEval} shows the high-confidence policy evaluation results for Enduro. Both the average returns over the posterior as well as the the high-confidence performance bounds ($\delta = 0.05$) demonstrate accurate predictions relative to the ground-truth performance. The No-Op policy results in the racecar slowly moving along the track and losing the race. This policy is accurately predicted as being much worse than the other evaluation policies. We also evaluated the Mean and MAP policies found by optimizing the Mean reward and MAP reward from the posterior obtained using Bayesian REX. We found that the learned posterior is able to capture that the MAP policy is more than twice as good as the evaluation policy D and that the Mean policy has performance somewhere between the performance of policies B and C. These results show that Bayesian REX has the potential to predict better-than-demonstrator performance \cite{brown2019drex}.

\begin{table}[t]
  \caption{Policy evaluation statistics for Enduro over the  return distribution from the learned posterior $P(R | D, \mathcal{P})$ compared with the ground truth returns using game scores. Policies A-D correspond to checkpoints of an RL policy partially trained on the ground-truth reward function and correspond to 25, 325, 800, and 1450 training updates to PPO. No-Op that always plays the no-op action, resulting in high mean predicted performance but low 95\%-confidence return (0.05-VaR). }
  \label{tab:enduroPolicyEval}
  \centering
  \vspace{0.1cm}
  \begin{tabular}{ccccc}
    \toprule
    & \multicolumn{2}{c}{Predicted}  & \multicolumn{2}{c}{Ground Truth} \\
 Policy & Mean & 0.05-VaR& Avg. & Length \\
 \midrule
A & 324.7 & 48.2  & 7.3 &  3322.4  \\
B & 328.9 & 52.0  & 26.0 & 3322.4  \\
C & 424.5 & 135.8  & 145.0 & 3389.0  \\
D & 526.2 & 192.9  & 199.8 & 3888.2  \\
Mean & 1206.9 & 547.5  & 496.7 & 7249.4  \\
MAP & 395.2 & 113.3  & 133.6 & 3355.7  \\
No-Op & 245.9 & -31.7  & 0.0 & 3322.0  \\
    \bottomrule
  \end{tabular}
\end{table}

Table~\ref{tab:seaquestPolicyEval} shows the results for high-confidence policy evaluation for Seaquest. The results show that high-confidence performance bounds are able to accurately predict that evaluation policies A and B are worse than C and D. The ground truth performance of policies C and D are too close and the mean performance over the posterior and 0.05-VaR bound on the posterior are not able to find any statistical difference between them. Interestingly the no-op policy has very high mean and 95\%-confidence lower bound, despite not scoring any points. However, as shown in the bottom half of Table~\ref{tab:seaquestPolicyEval}, adding one more ranked demonstration from a 3000 length segment of a no-op policy solves this problem. These results motivate a natural human-in-the-loop approach for safe imitation learning.

\begin{table}[t]
  \caption{Policy evaluation statistics for Seaquest over the  return distribution from the learned posterior $P(R | D, \mathcal{P})$ compared with the ground truth returns using game scores. Policies A-D correspond to checkpoints of an RL policy partially trained on the ground-truth reward function and correspond to 25, 325, 800, and 1450 training updates to PPO. No-Op always plays the no-op action, resulting in high mean predicted performance but low 0.05-quantile return (0.05-VaR). Results predict that No-Op is much better than it really is. However, simply adding a single ranked rollout from the No-Op policy and rerunning MCMC results in correct relative rankings with respect to the No-Op policy}
  \label{tab:seaquestPolicyEval}
  \centering
  \vspace{0.1cm}
  \begin{tabular}{ccccc}
    \toprule
    & \multicolumn{2}{c}{Predicted}  & \multicolumn{2}{c}{Ground Truth}  \\
 Policy & Mean & 0.05-VaR& Avg. & Length \\
 \midrule
A & 24.3 & 10.8  & 338.6 & 1077.8  \\
B & 53.6 & 24.1  & 827.2 & 2214.1  \\
C & 56.0 & 25.4  & 872.2 & 2248.5  \\
D & 55.8 & 25.3  & 887.6 & 2264.5  \\
No-Op & 2471.6 & 842.5  & 0.0 & 99994.0  \\
\midrule
\multicolumn{5}{c}{Results after adding one ranked demo from No-Op}\\
 \midrule
A & 0.5 & -0.5  & 338.6 & 1077.8  \\
B & 3.7 & 2.0  & 827.2 & 2214.1  \\
C & 3.8 & 2.1  & 872.2 & 2248.5  \\
D & 3.2 & 1.5  & 887.6 & 2264.5  \\
No-Op & -321.7 & -578.2  & 0.0 & 99994.0  \\
    \bottomrule
  \end{tabular}
\end{table}

Finally, Table~\ref{tab:spaceinvadersPolicyEval} shows the results for high-confidence policy evaluation for Space Invaders. The results show that using both the mean performance and 95\%-confidence lower bound are good indicators of ground truth performance for the evaluation polices. The No-Op policy for Space Invaders results in the agent getting hit by alien lasers early in the game. The learned reward function posterior correctly assigns low average performance and indicates high risk with a low 95\%-confidence lower bound on the expected return of the evaluation policy.

\begin{table}[t]
  \caption{Policy evaluation statistics for Space Invaders over the  return distribution from the learned posterior $P(R | D, \mathcal{P})$ compared with the ground truth returns using game scores. Policies A-D correspond to checkpoints of an RL policy partially trained on the ground-truth reward function and correspond to 25, 325, 800, and 1450 training updates to PPO. The mean and MAP policies are the results of PPO using the mean and MAP rewards, respectively. No-Op that always plays the no-op action, resulting in high mean predicted performance but low 0.05-quantile return (0.05-VaR).}
  \label{tab:spaceinvadersPolicyEval}
  \centering
  \vspace{0.1cm}
  \begin{tabular}{ccccc}
    \toprule
    & \multicolumn{2}{c}{Predicted}  & \multicolumn{2}{c}{Ground Truth}  \\
 Policy & Mean & 0.05-VaR& Avg. & Length \\
 \midrule
A & 45.1 & 20.6  & 195.3 & 550.1  \\
B & 108.9 & 48.7  & 436.0 & 725.7  \\
C & 148.7 & 63.6  & 575.2 & 870.6  \\
D & 150.5 & 63.8  & 598.2 & 848.2  \\
Mean & 417.4 & 171.7  & 1143.7 & 1885.7  \\
MAP & 360.2 & 145.0  & 928.0 & 1629.5  \\
NoOp & 18.8 & 3.8  & 0.0 & 504.0  \\

    \bottomrule
  \end{tabular}
\end{table}

\section{Different Evaluation Policies}
To test Bayesian REX on different learned policies we took a policy trained with RL on the ground truth reward function for Beam Rider, the MAP policy learned via Bayesian REX for Beam Rider, and a policy trained with an earlier version of Bayesian REX (trained without all of the auxiliary losses) that learned a novel reward hack where the policy repeatedly presses left and then right, enabling the agent's ship to stay in between two of the firing lanes of the enemies. The imitation learning reward hack allows the agent to live for a very long time. We took a 2000 step prefix of each policy and evaluated the expected and 5th perentile worst-case predicted returns for each policy. We found that Bayesian REX is able to accurately predict that the reward hacking policy is worse than both the RL policy and the policy optimizing the Bayesian REX reward. However, we found that the Bayesian REX policy, while not performing as well as the RL policy, was given higher expected return and a higher lower bound on performance than the RL policy. Results are shown in Table~\ref{tab:beamriderRewardHacking}.

\begin{table}[t]
  \caption{Beamrider policy evaluation for an RL policy trained on ground truth reward, an imitation learning policy, and a reward hacking policy that exploits a game hack to live for a long time by moving quickly back and forth.}
  \label{tab:beamriderRewardHacking}
  \centering
  \vspace{0.1cm}
  \begin{tabular}{ccccc}
    \toprule
    & \multicolumn{2}{c}{Predicted}  & \multicolumn{2}{c}{Ground Truth}  \\
 Policy & Mean & 0.05-VaR& Avg. & Length \\
 \midrule
RL & 36.7 & 19.5 & 2135.2 & 2000 \\
B-REX & 68.1 & 38.1 & 649.4 & 2000 \\
Hacking & 28.8 & 10.2 & 2.2 & 2000\\

    \bottomrule
  \end{tabular}
\end{table}

%We found that when we evaluated these policies for a fixed horizon that the posterior is able to appropriately detect that the reward hacking behavior is worse than the RL policy or the imitation learning policy found with Bayesian REX. 

\section{Human Demonstrations}
To investigate whether Bayesian REX is able to correctly rank human demonstrations, one of the authors provided demonstrations of a variety of different behaviors and then we took the latent embeddings of the demonstrations and used the posterior distribution to find high-confidence performance bounds for these different rollouts.

\subsection{Beamrider}
We generated four human demonstrations: (1) \textit{good}, a good demonstration that plays the game well, (2) \textit{bad}, a bad demonstration that seeks to play the game but does a poor job, (3) \textit{pessimal}, a demonstration that does not shoot enemies and seeks enemy bullets, and (4) \textit{adversarial} a demonstration that pretends to play the game by moving and shooting as much as possibly but tries to avoid actually shooting enemies. The results of high-confidence policy evaluation are shown in Table~\ref{tab:beamRiderHumanDemoPolicyEval_app}. The high-confidence bounds and average performance over the posterior correctly rank the behaviors. This provides evidence that the learned linear reward correctly rewards actually destroying aliens and avoiding getting shot, rather than just flying around and shooting.

\begin{table}[t]
  \caption{Beam Rider evaluation of a variety of human demonstrations.}
  \label{tab:beamRiderHumanDemoPolicyEval_app}
  \centering
  \vspace{0.1cm}
  \begin{tabular}{ccccc}
    \toprule
    & \multicolumn{2}{c}{Predicted}  & \multicolumn{2}{c}{Ground Truth}  \\
 Policy & Mean & 0.05-VaR& Avg. & Length \\
 \midrule
good & 12.4 & 5.8 & 1092 & 1000.0 \\
bad & 10.7 & 4.5 & 396 & 1000.0 \\
pessimal & 6.6 & 0.8 & 0 & 1000.0 \\
adversarial & 8.4 & 2.4 & 176 & 1000.0 \\
\bottomrule
  \end{tabular}
\end{table}

\subsection{Space Invaders}
For Space Invaders we demonstrated an even wider variety of behaviors to see how Bayesian REX would rank their relative performance. We evaluated the following policies: (1) \textit{good}, a demonstration that attempts to play the game as well as possible, (2) \textit{every other}, a demonstration that only shoots aliens in the 2nd and 4th columns, (3) \textit{flee}, a demonstration that did not shoot aliens, but tried to always be moving while avoiding enemy lasers, (4) \textit{hide}, a demonstration that does not shoot and hides behind on of the barriers to avoid enemy bullets,  (5) \textit{pessimal}, a policy that seeks enemy bullets while not shooting, (6) \textit{shoot shelters}, a demonstration that tries to destroy its own shelters by shooting at them, (7) \textit{hold 'n fire}, a demonstration where the player rapidly fires but does not move to avoid enemy lasers, and (8) \textit{miss}, a demonstration where the demonstrator tries to fire but not hit any aliens while avoiding enemy lasers. 

\begin{table}
  \caption{Space Invaders evaluation of a variety of human demonstrations.}
  \label{tab:spaceinvadersHumanDemoPolicyEval}
  \centering
  \vspace{0.1cm}
  \begin{tabular}{ccccc}
    \toprule
    & \multicolumn{2}{c}{Predicted}  & \multicolumn{2}{c}{Ground Truth}  \\
 Policy & Mean & 0.05-VaR& Avg. & Length \\
 \midrule
good & 198.3 & 89.2 & 515 & 1225.0 \\
every other & 56.2 & 25.9 & 315 & 728.0 \\
hold 'n fire & 44.3 & 18.6 & 210 & 638.0 \\
shoot shelters & 47.0 & 20.6 & 80 & 712.0 \\
flee & 45.1 & 19.8 & 0 & 722.0 \\
hide & 83.0 & 39.0 & 0 & 938.0 \\
miss & 66.0 & 29.9 & 0 & 867.0 \\
pessimal & 0.5 & -13.2 & 0 & 266.0 \\
\bottomrule
  \end{tabular}
\end{table}

Table~\ref{tab:spaceinvadersHumanDemoPolicyEval} shows the results of evaluating the different demonstrations. The good demonstration is clearly the best performing demonstration in terms of mean performance and 95\%-confidence lower bound on performance and the pessimal policy is correctly given the lowest performance lower bound. However, we found that the length of the demonstration appears to have a strong effect on the predicted performance for Space Invaders. Demonstrations such as hide and miss are able to live for a longer time than policies that actually hit aliens. This results in them having higher 0.05-quantile worst-case predicted performance and higher mean performance. 

To study this further we looked at only the first 600 timesteps of each policy, to remove any confounding by the length of the trajectory. The results are shown in Table~\ref{tab:spaceinvadersHumanDemoPolicyEval_truncated}. With a fixed length demonstration, Bayesian REX is able to correctly rank \textit{good}, \textit{every other}, and \textit{hold 'n fire} as the best demonstrations, despite evaluation policies that are deceptive.

\begin{table}
  \caption{Space Invaders evaluation of a variety of human demonstrations when considering only the first 6000 steps.}
  \label{tab:spaceinvadersHumanDemoPolicyEval_truncated}
  \centering
  \vspace{0.1cm}
  \begin{tabular}{ccccc}
    \toprule
    & \multicolumn{2}{c}{Predicted}  & \multicolumn{2}{c}{Ground Truth}  \\
 Policy & Mean & 0.05-VaR& Avg. & Length \\
 \midrule
good & 47.8 & 22.8 & 515 &600.0 \\
every other & 34.6 & 15.0 & 315 & 600.0 \\
hold 'n fire & 40.9 & 17.1 & 210 & 600.0 \\
shoot shelters & 33.0 & 13.3 & 80 & 600.0 \\
flee & 31.3 & 11.9 & 0 & 600.0 \\
hide & 32.4 & 13.8 & 0 & 600.0 \\
miss & 30.0 & 11.3 & 0 & 600.0 \\
\bottomrule
  \end{tabular}
\end{table}

\subsection{Enduro}
For Enduro we tested four different human demonstrations: (1) \textit{good} a demonstration that seeks to play the game well, (2) \textit{periodic} a demonstration that alternates between speeding up and passing cars and then slowing down and being passed, (3) \textit{neutral} a demonstration that stays right next to the last car in the race and doesn't try to pass or get passed, and (4) \textit{ram} a demonstration that tries to ram into as many cars while going fast. Table~\ref{tab:enduroHumanDemoPolicyEval_app} shows that Bayesian REX is able to accurately predict the performance and risk of each of these demonstrations and gives the highest (lowest 0.05-VaR) risk to the \textit{ram} demonstration and the least risk to the \textit{good} demonstration.

\begin{table}
  \caption{Enduro evaluation of a variety of human demonstrations.}
  \label{tab:enduroHumanDemoPolicyEval_app}
  \centering
  \vspace{0.1cm}
  \begin{tabular}{ccccc}
    \toprule
    & \multicolumn{2}{c}{Predicted}  & \multicolumn{2}{c}{Ground Truth}  \\
 Policy & Mean & 0.05-VaR& Avg. & Length \\
 \midrule
good & 246.7 & -113.2 & 177 & 3325.0 \\
periodic & 230.0 & -130.4 &  44 & 3325.0 \\
neutral & 190.8 & -160.6 & 0 & 3325.0 \\
ram & 148.4 & -214.3 & 0 & 3325.0 \\
\bottomrule
  \end{tabular}
\end{table}

\section{Comparison with Other Methods for Uncertainty Quantification}
Bayesian REX is only one possible method for measure uncertainty. Other popular methods for measuring epistemic uncertainty include using bootstrapping to create an ensemble of neural networks \cite{lakshminarayanan2017simple} and using dropout as an approximation of MCMC sampling \cite{gal2016dropout}. In this section we compare our fully Bayesian approach with these two approximations.

\subsection{T-REX Ensemble}
We used the same implementation used by \citet{browngoo2019trex}\footnote{https://github.com/hiwonjoon/icml2019-trex}, but trained an ensemble of five T-REX networks using the same training demonstrations but with randomized seeds so each network is intitialized differently and has a different training set of subsampled snippets from the full length ranked trajectories. To estimate the uncertainty over the return of a trajectory or policy we run the trajectory through each network to get a return estimate or run multiple rollouts of the policy through each member of the ensemble to get a distribution over returns. We used 100 rollouts for the evaluation policies.

\subsection{MC Dropout}
For the MC Dropout baseline we used the same base architecture as T-REX and Bayesian REX, except that we did not add additional auxiliary losses, but simply trained the base network to predict returns using dropout during training. For each training pair of trajectories we randomly sample a dropout mask on the last layer of weights. Because MC dropout is supposed to approximate a large ensemble, we kept the dropout mask consistent across each sampled preference pair such that the same portions of the network are dropped out for each frame of each trajectory and for both the more preferred and less preferred trajectories. Thus, for each training sample, consisting of a more and less preferred trajectory, we sample a random dropout mask and then apply this same mask across all states in both trajectories. 
To keep things as similar to Bayesian REX as possible, we used full trajectories with the same pairwise preferences used by Bayesian REX.

To estimate the posterior distribution over returns for a trajectory or policy we simply sampled 50 random masks on the last layer of weights. Thus, this method corresponds to the MC dropout equivalent of Bayesian REX where the latent state encoding is trained end-to-end via dropout rather than pre-trained and where the posterior distribution is estimated via randomly dropping out weights on the corresponding linear reward function. We applied these 50 random dropouts to each of 100 rollouts for each evaluation policy. We used a dropout probability of 0.5.

Table~\ref{tab:brex_ensemble_mcdropout} shows the results for running RL on the learned reward functions. The results show that Bayesian REX is superior or competitive with T-REX Ensemble and MC Dropout across all games except Beam Rider, where MC Dropout performs much better. 

\subsection{T-REX Ensemble High-Confidence Bounds}
Tables~\ref{tab:beamriderEvalEnsemble}--\ref{tab:spaceinvadersEvalEnsemble} show the results for evaluating different evaluation policies via high-confidence performance bounds. Table~\ref{tab:beamriderEvalEnsemble} shows that the ensemble has accurate expected returns, but that the 95\% confidence lower bounds are not informative and do not represent risk as accurately as Bayesian REX since policy D is viewed as much worse than policy A. Note that we normalized the predicted scores by calculating the average predicted return of each ensemble member for rollouts from policy A and then using this as a baseline for all other predictions of each ensemble member by subtracting off the average predicted return for policy A from return predictions of other policies. Tables~\ref{tab:breakoutEvalEnsemble} and \ref{tab:seaquestEvalEnsemble} show that the T-REX Ensemble can sometimes fail to produce meaningful predictions for the expectation or the 95\% worst-case bounds. Table~\ref{tab:enduroEvalEnsemble} and \ref{tab:spaceinvadersEvalEnsemble} show good predictions.

\begin{table*}[t]
  \caption{Comparison of policy performance when using a reward function learned by Bayesian REX, a T-REX ensemble, and a dropout version of Bayesian REX. The results show averages (standard deviations over 30 rollouts. Bayesian REX results in comparable or better performance across all games except Beam Rider.}
  \label{tab:brex_ensemble_mcdropout}
  \centering
    \vspace{0.1cm}
\begin{tabular}{ccccccc}
\toprule
 & \multicolumn{2}{c}{Bayesian REX} &\\
Game &  Mean & MAP & T-REX Ensemble & MC Dropout\\ 
\midrule 
Beam Rider & 5870.3 (1905.1)  & 5504.7 (2121.2)  & 5925.0 (2097.9)  & \textbf{7243.1} (2543.6)\\ 
Breakout & \textbf{393.1} (63.7)  & 390.7 (48.8)  & 253.7 (136.2)  & 52.6 (10.1)\\ 
Enduro & 135.0 (24.8)  & \textbf{487.7} (89.4)  & 281.5 (95.2)  & 111.8 (17.9)\\ 
Seaquest & 606.0 (37.6)  & \textbf{734.7} (41.9)  & 0.0 (0.0)  & 0.0 (0.0)\\ 
Space Invaders & 961.3 (392.3)  & 1118.8 (483.1)  & \textbf{1164.8} (546.3)  & 387.5 (166.3)\\ 
\bottomrule
\end{tabular}
\end{table*}

\begin{table}[t]
  \caption{Beam Rider T-REX ensemble.}
  \label{tab:beamriderEvalEnsemble}
  \centering
  \vspace{0.1cm}
  \begin{tabular}{ccccc}
    \toprule
    & \multicolumn{2}{c}{Predicted}  & \multicolumn{2}{c}{Ground Truth Avg.}  \\
 Policy & Mean & 0.05-VaR& Score & Length\\
 \midrule
A & 0.0 & -119.5 & 454.4 & 1372.6  \\
B & 76.1 & -63.7 & 774.8 & 1412.8  \\
C & 201.3 & -173.8 & 1791.8 & 2389.9  \\
D & 282.0 & -304.0 & 2664.5 & 2965.0  \\
Mean & 956.9 & -2294.8 & 5736.8 & 9495.6  \\
MAP & 1095.7 & -2743.4 & 5283.0 & 11033.4  \\
No-Op & -1000.0 & -5643.7 & 0.0 & 99,994.0  \\
    \bottomrule
  \end{tabular}
\end{table}

\begin{table}[t]
  \caption{Breakout T-REX ensemble.}
  \label{tab:breakoutEvalEnsemble}
  \centering
  \vspace{0.1cm}
  \begin{tabular}{ccccc}
    \toprule
    & \multicolumn{2}{c}{Predicted}  & \multicolumn{2}{c}{Ground Truth Avg.}  \\
 Policy & Mean & 0.05-VaR& Score & Length\\
 \midrule
A & 0.0 & -506.3 & 1.8 & 202.7  \\
B & -305.8 & -1509.1 & 16.1 & 608.4  \\
C & -241.1 & -1780.7 & 24.8 & 849.3  \\
D & -57.9 & -2140.0 & 42.7 & 1020.8  \\
Mean & -5389.5 & -867.4 & 388.9 & 13762.1  \\
MAP & -3168.5 & -1066.5 & 401.0 & 8780.0  \\
No-Op & -39338.4 & -95987.2 & 0.0 & 99,994.0  \\

    \bottomrule
  \end{tabular}
\end{table}

\begin{table}[t]
  \caption{Enduro T-REX ensemble.}
  \label{tab:enduroEvalEnsemble}
  \centering
  \vspace{0.1cm}
  \begin{tabular}{ccccc}
    \toprule
    & \multicolumn{2}{c}{Predicted}  & \multicolumn{2}{c}{Ground Truth Avg.}  \\
 Policy & Mean & 0.05-VaR& Score & Length\\
 \midrule
A & -0.0 & -137.7 & 9.4 & 3322.4  \\
B & 23.6 & -157.4 & 23.2 & 3322.4  \\
C & 485.4 & 232.2 & 145.6 & 2289.0  \\
D & 1081.1 & 270.3 & 214.2 & 3888.2  \\
Mean & 3408.9 & 908.0 & 496.7 & 7249.4  \\
MAP & 23.2 & -1854.1 & 133.6 & 3355.7  \\
No-Op & -1618.1 & -3875.9 & 0.0 & 3322.0  \\    \bottomrule
  \end{tabular}
\end{table}

\begin{table}[t]
  \caption{Seaquest T-REX ensemble.}
  \label{tab:seaquestEvalEnsemble}
  \centering
  \vspace{0.1cm}
  \begin{tabular}{ccccc}
    \toprule
    & \multicolumn{2}{c}{Predicted}  & \multicolumn{2}{c}{Ground Truth Avg.}  \\
 Policy & Mean & 0.05-VaR& Score & Length\\
 \midrule
A & 0.0 & -320.9 & 321.0 & 1077.8  \\
B & -425.2 & -889.3 & 826.6 & 2214.1  \\
C & -336.7 & -784.3 & 863.4 & 2248.5  \\
D & -386.3 & -837.3 & 884.4 & 2264.5  \\
Mean & -1013.1 & -2621.8 & 721.8 & 2221.7  \\
MAP & -636.8 & -1820.1 & 607.4 & 2247.2  \\
No-Op & -19817.9 & -28209.7 & 0.0 & 99,994.0  \\  \bottomrule
  \end{tabular}
\end{table}

\begin{table}[t]
  \caption{Space Invaders T-REX ensemble.}
  \label{tab:spaceinvadersEvalEnsemble}
  \centering
  \vspace{0.1cm}
  \begin{tabular}{ccccc}
    \toprule
    & \multicolumn{2}{c}{Predicted}  & \multicolumn{2}{c}{Ground Truth Avg.}  \\
 Policy & Mean & 0.05-VaR& Score & Length\\
 \midrule
A & 0.0 & -136.8 & 159.4 & 550.1  \\
B & 257.3 & -47.9 & 425.0 & 725.7  \\
C & 446.5 & -6.0 & 553.1 & 870.6  \\
D & 443.3 & 9.0 & 591.5 & 848.2  \\
Mean & 1105.6 & -392.4 & 1143.7 & 1885.7 \\
MAP & 989.0 & -387.2 & 928.0 & 1629.5  \\
No-Op & -211.9 & -311.9 & 0.0 & 504.0  \\
 \bottomrule
  \end{tabular}
\end{table}

\subsection{MC Dropout Results}
Tables~\ref{tab:beamriderMCdropoutEval}--\ref{tab:spaveinvadersMCdropoutEval} show the results for high-confidence bounds for MC Dropout. Tables~\ref{tab:beamriderMCdropoutEval} and \ref{tab:spaveinvadersMCdropoutEval} show that MC Dropout is able to accurately predict high risk for the Beam Rider and Space Invaders No-Op policies. However, table~\ref{tab:breakoutMCdropoutEval} \ref{tab:enduroMCdropoutEval}, and \ref{tab:seaquestMCdropoutEval} show that MC Dropout often fails to predict that the No-Op policy has high risk. Recent work has shown that MC Dropout is not a principled Bayesian approximation since the distribution obtained from MC Dropout does not concentrate in the limit as the number of data samples goes to infinity and thus does not necessarily measure the kind of epistemic risk we are interested in \cite{osband2016risk}. Thus, while MC Dropout does not perform full Bayesian inference like Bayesian REX, it appears to work sometimes in practice. Future work should examine more sophisticated applications of dropout to uncertainty estimation that seek to solve the theoretical and practical problems with vanilla MC Dropout \cite{hron2018variational}.   

\begin{table}[t]
  \caption{Beam Rider MC Dropout.}
  \label{tab:beamriderMCdropoutEval}
  \centering
  \vspace{0.1cm}
  \begin{tabular}{ccccc}
    \toprule
    & \multicolumn{2}{c}{Predicted}  & \multicolumn{2}{c}{Ground Truth Avg.}  \\
 Policy & Mean & 0.05-VaR& Score & Length\\
 \midrule
A & 20.9 & -1.5 & 454.4 & 1372.6  \\
B & 27.9 & 2.3 & 774.8 & 1412.8  \\
C & 48.7 & 8.3 & 1791.8 & 2389.9  \\
D & 63.5 & 11.0 & 2664.5 & 2965.0  \\
Mean & 218.2 & -89.2 & 5736.8 & 1380  \\
MAP & 211.2 & -148.7 & 5283.0 & 708  \\
No-Op & 171.2 & -3385.7 & 0.0 & 99,994.0  \\
 \bottomrule
  \end{tabular}
\end{table}

\begin{table}[t]
  \caption{Breakout MC Dropout.}
  \label{tab:breakoutMCdropoutEval}
  \centering
  \vspace{0.1cm}
  \begin{tabular}{ccccc}
    \toprule
    & \multicolumn{2}{c}{Predicted}  & \multicolumn{2}{c}{Ground Truth Avg.}  \\
 Policy & Mean & 0.05-VaR& Score & Length\\
 \midrule
A & 10.8 & 5.2 & 1.8 & 202.7  \\
B & 33.1 & 17.7 & 16.1 & 608.4  \\
C & 43.5 & 24.1 & 24.8 & 849.3  \\
D & 56.0 & 28.5 & 42.7 & 1020.8  \\
Mean & 822.9 & 77.3 & 388.9 & 13762.1  \\
MAP & 519.7 & 73.8 & 401.0 & 8780.0  \\
No-Op & 6050.7 & 3912.4 & 0.0 & 99,994.0 \\
 \bottomrule
  \end{tabular}
\end{table}

\begin{table}[t]
  \caption{Enduro MC Dropout.}
  \label{tab:enduroMCdropoutEval}
  \centering
  \vspace{0.1cm}
  \begin{tabular}{ccccc}
    \toprule
    & \multicolumn{2}{c}{Predicted}  & \multicolumn{2}{c}{Ground Truth Avg.}  \\
 Policy & Mean & 0.05-VaR& Score & Length\\
 \midrule
A & 541.7 & 398.0 & 7.3 & 3322.4  \\
B & 543.6 & 401.0 & 26.4 & 3322.4  \\
C & 556.7 & 409.3 & 142.5 & 3389.0  \\
D & 663.3 & 422.3 & 200.3 & 3888.2  \\
Mean & 2473.0 & 1701.7 & 496.7 & 7249.4  \\
MAP & 1097.3 & 799.5 & 133.6 & 3355.7  \\
No-Op & 1084.1 & 849.8 & 0.0 & 3322.0  \\
 \bottomrule
  \end{tabular}
\end{table}

\begin{table}[t]
  \caption{Seaquest MC Dropout.}
  \label{tab:seaquestMCdropoutEval}
  \centering
  \vspace{0.1cm}
  \begin{tabular}{ccccc}
    \toprule
    & \multicolumn{2}{c}{Predicted}  & \multicolumn{2}{c}{Ground Truth Avg.}  \\
 Policy & Mean & 0.05-VaR& Score & Length\\
 \midrule
A & 98.9 & 49.5 & 321.0 & 1077.8  \\
B & 258.8 & 194.8 & 826.6 & 2214.1  \\
C & 277.7 & 213.2 & 863.4 & 2248.5  \\
D & 279.6 & 214.2 & 884.4 & 2264.5  \\
Mean & 375.6 & 272.8 & 721.8 & 2221.7  \\
MAP & 426.3 & 319.8 & 607.4 & 2247.2  \\
No-Op & 16211.1 & 10478.5 & 0.0 & 99,994.0  \\  \bottomrule
  \end{tabular}
\end{table}

\begin{table}[t]
  \caption{Space Invaders MC Dropout.}
  \label{tab:spaveinvadersMCdropoutEval}
  \centering
  \vspace{0.1cm}
  \begin{tabular}{ccccc}
    \toprule
    & \multicolumn{2}{c}{Predicted}  & \multicolumn{2}{c}{Ground Truth Avg.}  \\
 Policy & Mean & 0.05-VaR& Score & Length\\
 \midrule
A & 10.6 & 0.8 & 195.3 & 550.1  \\
B & 22.3 & 8.8 & 434.9 & 725.7  \\
C & 26.7 & 9.8 & 535.3 & 870.6  \\
D & 28.9 & 15.6 & 620.9 & 848.2  \\
Mean & 125.9 & 54.4 & 1143.7 & 848.2  \\
MAP & 110.6 & 52.5 & 928.0 & 1885.7  \\
No-Op & 8.4 & -8.6 & 0.0 & 504.0  \\
  \bottomrule
  \end{tabular}
\end{table}
% \textbf{\emph{Do not put content after the references.}}
% %
% Put anything that you might normally include after the references in a separate
% supplementary file.

% We recommend that you build supplementary material in a separate document.
% If you must create one PDF and cut it up, please be careful to use a tool that
% doesn't alter the margins, and that doesn't aggressively rewrite the PDF file.
% pdftk usually works fine. 

% \textbf{Please do not use Apple's preview to cut off supplementary material.} In
% previous years it has altered margins, and created headaches at the camera-ready
% stage. 
%%%%%%%%%%%%%%%%%%%%%%%%%%%%%%%%%%%%%%%%%%%%%%%%%%%%%%%%%%%%%%%%%%%%%%%%%%%%%%%
%%%%%%%%%%%%%%%%%%%%%%%%%%%%%%%%%%%%%%%%%%%%%%%%%%%%%%%%%%%%%%%%%%%%%%%%%%%%%%%

\end{document}